\pdfoutput=1
\PassOptionsToPackage{prologue,dvipsnames}{xcolor}
\documentclass[11pt]{article}

\usepackage[]{acl}

\usepackage{times}
\usepackage{latexsym}

\usepackage[T1]{fontenc}
\usepackage[utf8]{inputenc}

\usepackage{microtype}

\usepackage{inconsolata}

\usepackage{amsfonts}
\usepackage{xcolor}
\usepackage{anyfontsize}

\usepackage{graphicx}
\usepackage{multirow}
\usepackage{amsmath}
\usepackage{cleveref}
\usepackage{hyperref}
\usepackage{lipsum}
\usepackage{caption}
\usepackage{subcaption}
\usepackage{algorithm}
\usepackage{algpseudocode}
\usepackage{colortbl}
\usepackage{enumitem}
\usepackage[normalem]{ulem}
\useunder{\uline}{\ul}{}



\usepackage{todonotes}
\definecolor{taa-color}{rgb}{0,0,1}
\definecolor{HIGHLIGHT-COLOR}{rgb}{0.9, 0.1, 0.1}

\usepackage{framed} % or, "mdframed"
\usepackage[framed]{ntheorem}
\newframedtheorem{frm-thm}{Performance Metric}

\title{XForecast: Evaluating Natural Language Explanations for Time Series Forecasting}

\author{Taha Aksu\thanks{Taha Aksu is the corresponding author. Email: iaksu@salesforce.com}, Chenghao Liu, Amrita Saha, Sarah Tan, Caiming Xiong, Doyen Sahoo\\
Salesforce AI Research, \\
\texttt{\{iaksu,chenghao.liu,amrita.saha,sarah.tan,cxiong,dsahoo\}@salesforce.com} \\
}

\begin{document}
\maketitle
\begin{abstract}
Time series forecasting aids decision-making, especially for stakeholders who rely on accurate predictions, making it very important to understand and explain these models to ensure informed decisions. Traditional explainable AI (XAI) methods, which underline feature or temporal importance, often require expert knowledge. In contrast, natural language explanations (NLEs) are more accessible to laypeople. However, evaluating forecast NLEs is difficult due to the complex causal relationships in time series data. To address this, we introduce two new performance metrics based on simulatability, assessing how well a human surrogate can predict model forecasts using the explanations. Experiments show these metrics differentiate good from poor explanations and align with human judgments. Utilizing these metrics, we further evaluate the ability of state-of-the-art large language models (LLMs) to generate explanations for time series data, finding that numerical reasoning, rather than model size, is the main factor influencing explanation quality.

\end{abstract}

\section{Introduction}
Time series forecasting is essential in various fields: in healthcare, it assists in early triage assessment~\cite{Morid2021TimeSP}; in finance, it helps detect investment opportunities~\cite{Sezer2019FinancialTS, Leung2020FinancialTS}; in human resources, it aids workforce planning~\cite{10467493}; and in energy, it promotes efficient consumption~\cite{DEB2017902}. For decision-makers to rely on forecasts, trust and transparency are essential, necessitating explanations. Explainable AI (XAI) is a field dedicated to addressing this challenge for general machine learning applications.

\begin{figure}
    \centering
    \includegraphics[width = 0.8\linewidth]{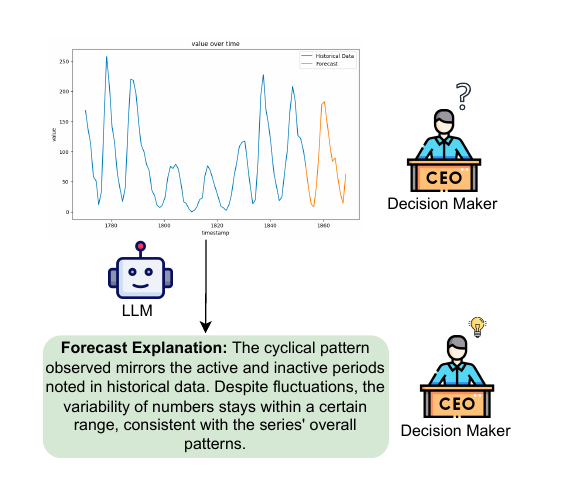}
    \caption{Example natural language explanation (NLE) for a time series forecast. While the raw forecast might be challenging for a layperson to interpret, the NLE provided by the LLM helps clarify the causal relationship.}
    \label{fig:intro_fig}
\end{figure}
XAI has also emerged in forecasting, where explanations typically use feature importance or saliency maps~\cite{ShapTimeZhang, Pan2020SeriesST, Crabbe2021ExplainingTS, Raykar2023TsSHAPRM, Rajapaksha2021LoMEFAF}. However, these explanations require technical expertise, making them suitable for AI engineers rather than end users~\cite{Fok2023InSO, interpret_tools, Lakkaraju2022RethinkingEA, Mavrepis2024XAIFA, Singh2024RethinkingII, Du2018TechniquesFI}. In contrast, natural language explanations (NLEs) are more user-friendly and can explain predictions from black-box models to a broader audience~\cite{Lakkaraju2022RethinkingEA, Mavrepis2024XAIFA}, see~\Cref{fig:intro_fig} for an example.

Despite its usefulness, evaluating natural language explanations for forecasts is challenging due to the unclear causal relationship between the historical and forecast windows of time series data. Unlike structured data like tables, time series data requires further processing to uncover patterns~\cite{Sharma2021T3DN}. This complicates distilling the technical depth of forecasting into concise NLEs and verifying their correctness.

To address this challenge, we propose two simulatability-based metrics: direct and synthetic simulatability. Direct simulatability measures how well a human can predict the black-box model’s outputs from explanations of the original time series. Synthetic simulatability tests how well the explanation generalizes to new, generated time series, ensuring it builds a mental model of the forecaster rather than overfitting to specific cases. To automate evaluation, we use a large language model (LLM) as a human surrogate following prior work~\cite{Aher2022UsingLL, Argyle2022OutOO, He2023AnnoLLMML, Gilardi2023ChatGPTOC}. To address the lack of annotated datasets of forecast-explanation pairs, we design a baseline explainer inspired by \citet{Warner1998} and conduct experiments to demonstrate the effectiveness of our proposed metrics. Our contributions are summarized as follows:
\begin{itemize}[leftmargin=*]
\item \textbf{New performance metrics:} We introduce two complementary simulatability-based performance metrics. Sanity checks and qualitative examples demonstrate their ability to distinguish between good and poor explanations. A two-part human study shows strong agreement between these metrics and human evaluations (Cohen's Kappa of 0.42 and 0.58), supporting their effectiveness in assessing explanations based on their usefulness to humans.
\item \textbf{Evaluating SoTA LLMs}: We then use these metrics to rank explanations generated by various LLMs. Our findings indicate that while LLM size impacts results, numerical reasoning capability is more crucial.
\end{itemize}

\section{Related Work}

\begin{figure*}[!htbp]
    \centering
    \includegraphics[width=\linewidth]{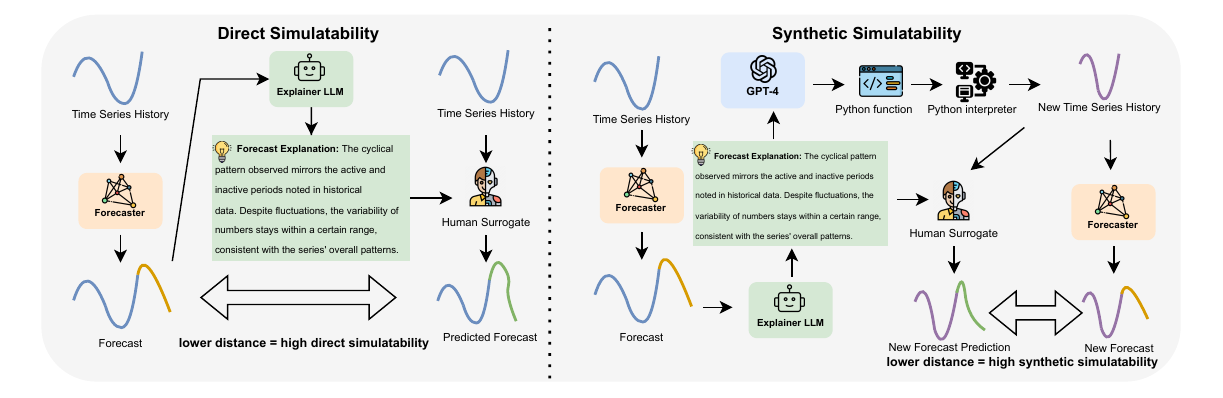}
    \caption{Depiction of newly proposed metrics for evaluating explanations of black box forecasting models. The left-hand side depicts the direct simulatability metric, which quantifies the distance between the ground truth and the simulated forecast on the original input. The right-hand side depicts the synthetic simulatability metric, which conducts simulations on a newly generated time series.}
    \label{fig:metrics}
\end{figure*}

\paragraph{Natural Language Explanations}
Pre-LLM XAI techniques mainly used structural explanations like saliency maps or feature importances due to the difficulty of generating text explanations. However, end users prefer text-based explanations. For example, \citet{Lakkaraju2022RethinkingEA} found that decision-makers such as doctors prefer natural language explanations, wanting to treat ML models as "another colleague." Similarly, \citet{Mavrepis2024XAIFA} found that over 80\% of participants preferred ChatGPT-refined XAI outputs in natural language. \citet{Du2018TechniquesFI} noted that many interpretable ML explanations are based on researchers' intuition rather than user needs, making them suitable for developers but not for lay users, and suggested conversational explanations as a more user-friendly alternative. Likewise, \citet{Singh2024RethinkingII} argued that LLMs can replace early XAI model interfaces, like saliency maps, enabling users to make targeted queries towards LLM-generated explanations. 
% In addition to the challenge of generating text explanations, another significant issue is their evaluation.

\paragraph{Evaluation of XAI Techniques}
Several techniques evaluate XAI methods, with plausibility and faithfulness being the most common. Plausibility measures how convincing the interpretation is to humans, while faithfulness assesses how accurately it reflects the model's true reasoning~\cite{jacovi-goldberg-2020-towards}. However, these metrics alone are insufficient. For instance, \citet{Fok2023InSO} reviewed many studies and concluded that explanations are only useful if they help human decision-makers verify AI predictions. \citet{interpret_tools} found that data scientists often over-trust and misuse interpretability tools, raising concerns about the quality of explanations provided to executives. Another important metric is simulatability, where humans must correctly simulate the model's output based on the explanation and input~\cite{DoshiVelez2017TowardsAR, chandrasekaran-etal-2018-explanations, hase-bansal-2020-evaluating}.~\citet{Chen2023DoME} applied simulatability to explain LLMs answers to text questions. They describe simulatability as a specific form of faithfulness, requiring human judgment rather than arbitrary black-box models. Since not every explanation is interpretable by humans, faithfulness does not imply simulatability. This concept guided the design of our proposed metrics as we adapted similar concepts to the time series modality.

\paragraph{XAI for Time Series Forecasting}
Various techniques are employed to explain forecast predictions. Shapley values are commonly used to explain the causal relationship between a model's features and its output. \citet{ShapTimeZhang} proposed a general XAI approach based on Shapley values that explores explanations in the temporal dimension and feeds these back to the forecasting model to improve performance. \citet{Raykar2023TsSHAPRM} introduced TsSHAP, which uses the TreeSHAP algorithm on a surrogate model to explain forecasts in terms of user-defined interpretable features. Building on LIME~\cite{Ribeiro2016WhySI}, \citet{Rajapaksha2021LoMEFAF} provide local model-agnostic interpretations by training simple surrogate models on samples within a neighborhood of the time series being explained. \citet{Pan2020SeriesST} and \citet{Crabbe2021ExplainingTS} convert multivariate time series into images to find saliency maps reflecting the importance of each temporal range across different features. While saliency maps and interpretable features are useful for domain experts or AI engineers, they are less intuitive for stakeholders.

\paragraph{This Work}
Unlike previous methods using feature importance or saliency matrices, we focus on natural language explanations (NLEs) and more importantly their evaluation. NLEs distill the complex causal relationships in time series into natural language. Our evaluation metrics draw from the concept of simulatability in XAI, assessing how well explanations help users form mental models of forecasting methods.

\section{New Evaluation Metrics}
\label{sec:metrics}

\subsection{Problem Definition}
\paragraph{Time Series Forecasting}
Time series (TS) data consists of a sequence of numerical values sampled at regular time intervals, denoted as $X^{(i)} = \{x_1^{(i)}, x_2^{(i)}, ..., x_{T}^{(i)}\}$, where $i\in \mathbb{Z}$ represents the number of variates, $T\in \mathbb{Z}$ it the number of timestamps, and $x_t^{(i)} \in \mathbb{R}$ is the value of $i_{th}$ variate at timestamp $t$. In this paper we focus exclusively on univariate time series, meaning $i=1$. Finally the objective of forecasting is to model the conditional distribution $P(x_{t:T} | x_{1:t-1})$. 
\paragraph{Natural Language Explanations}
For a given univariate time series data of length $t$, denoted as $H = \{h_1, h_2, ... h_t\}$, and the forecast prediction for the next $k$ time stamps, $F = \{f_1, f_2, ... f_k\}$, an explainer $E$ aims to generate a natural language explanation (NLE) that explains the causal relationship from $H$ to $F$. 
Thus given the triplet $\{ H, F, \text{NLE}\}$, our task is to evaluate the usefulness of NLE. 
We propose two complementary evaluation methods inspired by the simulatability metric from the XAI field~\cite{DoshiVelez2017TowardsAR}.
The first method is direct simulatability, which quantifies simulation on the original $\{ H, F, \text{NLE}\}$ triplet. However, relying solely on this may be misleading, as the explanation could be overly tailored to the specific example. Therefore, we also introduce synthetic simulatability, which evaluates how well the NLE generalizes by testing it on synthetic inputs, generating a new triplet ${H', F', \text{NLE}}$. Direct simulatability offers confidence in the explanation as it is directly tied to the original data, while synthetic simulatability ensures generalizability by testing the explanation on a loosely related ${H', F'}$ pair.

Evaluating time series forecasting with human subjects is costly and slow, so we use GPT-4 as a human surrogate for automated, end-to-end evaluation~\cite{Aher2022UsingLL, Argyle2022OutOO, He2023AnnoLLMML, Gilardi2023ChatGPTOC}. This approach reduces costs and logistical difficulties~\cite{Zheng2023JudgingLW, Chiang2023CanLL, cheng-etal-2023-gpt}. Specifically, we follow the procedure from LLMTime~\cite{gruver2023llmtime}, forecasting the next steps for a given time series by appending the generated explanation along with the historical context to the LLM prompt.

\subsection{Direct Simulatability}
\label{subsec:dir_sim}
\begin{algorithm}
\small
\caption{Evaluate Direct Simulatability}
\begin{algorithmic}[1]
\Require $H$: Time series data
\Require $\textit{FM}$: Forecasting model
\Require $\textit{Explainer}$: Forecast explainer
\Require $\textit{H.S.}$: Human surrogate to simulate the forecast
\Ensure $DS$: Direct simulatability score
\State $F \gets \textit{FM}(H) $ \Comment{Generate forecast}
\State $NLE \gets \textit{Explainer}(H, F)$ \Comment{Generate explanation}
\State $F' \gets \textit{H.S.}(H, NLE)$ \Comment{Human surrogate simulates forecast using $NLE$}
\State $DS \gets \textit{Distance}(F, F')$ \Comment{Calculate the distance between $F$ and $F'$}
\end{algorithmic}
\label{alg:dir_sim}
\end{algorithm}
% \begin{equation}
% \label{eq1}
% H + F  \rightarrow Explainer[LLM] \rightarrow NLE \\

% H + NLE \rightarrow Human Judge \rightarrow F' \\

% Distance(F,F') \rightarrow DS \\
% \end{equation}
Our first metric is based on the simulatability concept from XAI, which suggests that an explanation should help users form a mental model of the forecaster. We adapt this idea into a simulatability evaluation pipeline for time series forecasting (\textit{cf.} left side of~\Cref{fig:metrics}, and~\Cref{alg:dir_sim}). First, the black-box model generates a forecast, then the explainer provides an explanation, and finally, the human surrogate predicts the forecast using the explanation. We measure the distance between the actual and predicted forecasts—the better the explanation, the smaller the distance.

\subsection{Synthetic Simulatability}
\label{subsec:indir_sim}

\begin{figure}[htbp]
     \captionsetup[subfigure]{labelformat=empty} % This removes the (a) label
    \begin{subfigure}[t]{0.45\textwidth}
        \fbox{
        \begin{minipage}{\textwidth}
        \includegraphics[width=\textwidth]{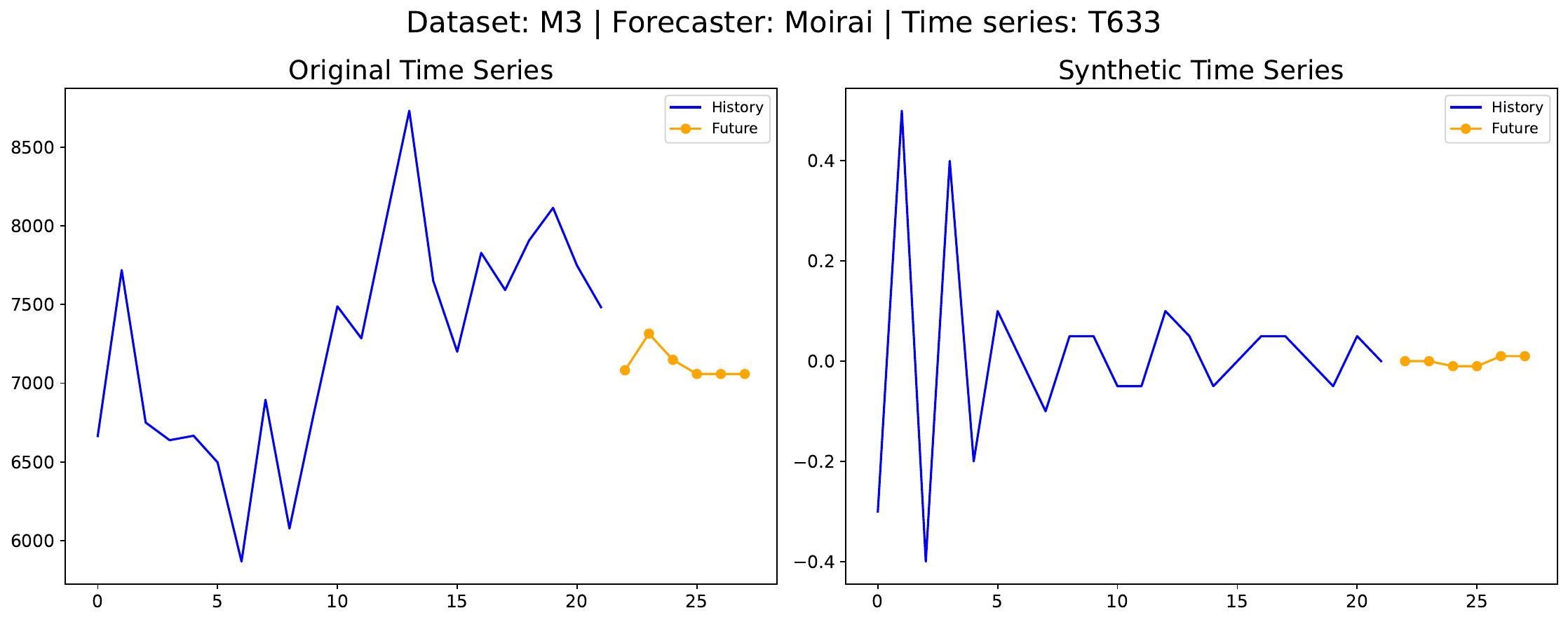}
        \caption{\textbf{Forecast Explanation:} The forecast suggests a stabilization in the time series, with an initial slight increase followed by a leveling off in the later time steps. This pattern of stabilization, indicated by the repeating value at the end of the forecast period, implies that the series may be reaching an equilibrium point following its previous volatility. While short-term fluctuations are present, the forecast does not anticipate a return to the higher variability seen earlier in the series, nor does it suggest a continuation of any significant upward or downward trend.}
        \end{minipage}
        }
        % \label{fig:synth_ex1}
    \end{subfigure}
\caption{The original and synthetic time series history and forecast pairs generated using the~\Cref{fig:metrics} pipeline (right) both align with the explanation, though they differ in scale and noise. The explanation is generated by the explainer using the original history and forecast.}
\label{fig:main_synth_example}
\end{figure}

A limitation of direct simulatability is that it only evaluates how well an explanation simulates the original time series. However, a useful explanation should help form a mental model that generalizes beyond a single series. To address this, we introduce synthetic simulatability, which assesses how well the explanation generalizes to new time series following the same reasoning.

\begin{algorithm}
\small
\caption{Evaluate Synthetic Simulatability}
\begin{algorithmic}[1]
\Require $H$: Time series data
\Require $\textit{FM}$: Forecasting model
\Require $\textit{Explainer}$: Forecast explainer
\Require $\textit{H.S.}$: Human surrogate to simulate the forecast
\Require $\textit{LLM}$: Large language model to generate new time series $NLE$
\Require $\textit{PI}$: Python interpreter
\Ensure $IS$: Synthetic simulatability score
\State $F \gets \textit{FM}(H)$ \Comment{Generate forecast}
\State $NLE \gets \textit{Explainer}(H, F)$ \Comment{Generate explanation}
\State $PF\gets \textit{LLM}(NLE)$ \Comment{Generate python function}
\State $H_{new} \gets \textit{PI}(PF)$ \Comment{Run python function to generate new time series}
\State $F_{new} \gets \textit{FM}(H_{new})$ \Comment{Generate new forecast}
\State $F_{new}' \gets \textit{H.S.}(H_{new}, NLE)$ \Comment{Human surrogate simulates new forecast using $NLE$}
\State $IS \gets \textit{Distance}(F_{new}, F_{new}')$ \Comment{Calculate the distance between $F_{new}$ and $F_{new}'$}
\end{algorithmic}
\label{alg:indir_sim}
\end{algorithm}

Unlike \citet{Chen2023DoME}, who use the same LLM to generate questions and explanations, our task is to generate time series data with explanations from a separate system. While their input is text-based, we generate numerical sequences, which is challenging for an LLM. To address this, we follow \citet{Merrill2024LanguageMS} by using GPT-4 to generate code that simulates time series data based on the explanations. This approach allows us to create new time series aligned with the explanations for synthetic simulatability evaluation.

Synthetic simulation pipeline (\textit{cf.} right side of~\Cref{fig:metrics}, and \Cref{alg:indir_sim}) starts with the black-box model generating a forecast and the explainer producing an explanation. GPT-4 then generates a new time series from this explanation using NL-to-code generation, where it creates a Python function to simulate the series (\textit{cf.} \Cref{app:eval_prompts} for detailed prompts and examples). The black-box model forecasts this new series, and the human surrogate predicts the forecast using the original explanation. We compare the model's forecast with the surrogate's prediction, and the distance between them quantifies synthetic simulatability: the smaller the distance, the better the explanation.

We present a sample synthetically generated time series in~\Cref{fig:main_synth_example}. Both the original and synthetic series align well with the explanation—e.g., both forecasts show stabilization without trends, and both historical contexts display high volatility with short-term fluctuations. However, their scales and noise levels differ significantly, so an explanation overfitting the original series would not aid in predicting the synthetic forecast. Additional examples are available in~\Cref{app:synth_examples},~\Cref{fig:synth_exs}.

\section{Baseline Explainer Design}
\label{sec:baseline}
To thoroughly test our evaluation metrics, we design a baseline explainer for generating NLEs for forecasts, as no existing baselines are available. For this baseline we refer to early work by \citet{Warner1998}, which offers techniques for interpreting time series data without requiring specialized knowledge. According to ~\citet{Warner1998-ch1, Warner1998-ch7} the key steps for explaining time series are: $i$)~\textbf{Statistics:} Screen the data to assess distribution, outliers, and relevant characteristics. $ii$)~\textbf{Trend:} Analyze linear trends to determine how much variance they account for. $iii$)~\textbf{Seasonality:} Look for cyclic patterns in the data. $iv$)~\textbf{Cycle Inconsistencies:} Describe changes in cycle irregularities, such as variations in peak amplitude over time.
% \begin{itemize}[leftmargin=*]
% \item \textbf{Statistics:} Start with data screening using usual methods to assess distribution properties, outliers, and other relevant characteristics.
% \item \textbf{Trend:} Perform a linear trend analysis to determine what percentage of the variance in the time series is accounted for by trends.
% \item \textbf{Seasonality:} Check for evidence of any cyclic patterns in the data.
% \item \textbf{Cycle Inconsistencies:} Describe the changes in irregularities in the cycles, \textit{i.e.}, how the amplitude or height of peaks changes over time.
% \end{itemize}
% \ta{Below paragraphs can go to appendix to save some space.}
Following the steps listed above, our pipeline first extracts characteristics using statistical methods. We then iteratively prompt the LLM to generate a summary of the time series, the forecast, and the relationship between the two. For detailed prompts and specific examples, refer to~\Cref{app:baseline_details}. 

\begin{figure}
    \centering
    \includegraphics[width=0.95\linewidth]{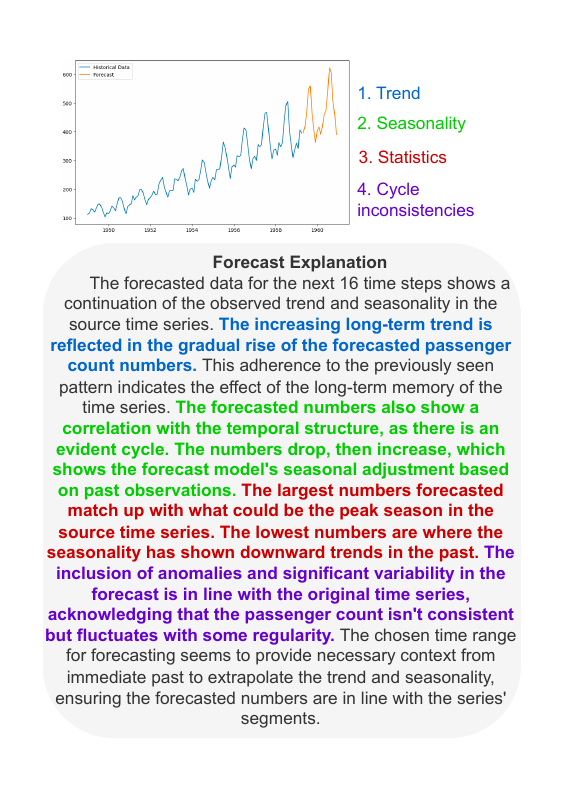}
    \caption{Sample forecasting explanation generated by our pipeline. The colored snippets refer to the 4 salient points by \citet{Warner1998} when explaining time series.}
    \label{fig:exp_sample}
\end{figure}

Given the length of time series data, we apply these steps to smaller segments and prompt the LLM to explain each separately, then aggregate them into a final explanation. Following~\citet{Sharma2021T3DN}, our pipeline first segments the time series based on slope changes. We then: $i)$ Perform quantitative analysis on each segment, calculating and formatting trend, seasonality, mean, and standard deviation into a templated summary; $ii)$ Concatenate the segment analyses and prompt the LLM to generate a comprehensive analysis of the full time series; $iii)$ Provide the LLM with the historical data, black-box forecast, and comprehensive analysis to generate a short report interpreting the forecast. See \textit{cf.}~\Cref{fig:exp_sample} for a sample explanation generated by this pipeline.

\section{Experiments and Analysis}
\label{sec:exps}

\subsection{Experimental setup}
\label{subsec:exp_setup}
\paragraph{Datasets}

We collected time series data from three datasets in the Monash Repository~\cite{godahewa2021monash}: Tourism, M3, and M1. To ensure the effectiveness of our metrics, the backbone LLM must perform reasonably well in forecasting. We focused on yearly frequencies due to shorter sequence lengths, as LLM forecasting performance declines with longer sequences~\cite{Fons2024EvaluatingLL}. As LLMs improve in time series reasoning, our metrics can be applied to any frequency. Exploration of other frequencies is left for future work.

\begin{table*}[ht]
\tiny
\centering
\begin{tabular}{|c|l|rrrrr|rrrrr|}
\hline
\multicolumn{1}{|l|}{}         &                & \multicolumn{5}{c|}{\textbf{Direct Simulatability}}                                                                                                                                                     & \multicolumn{5}{c|}{\textbf{Synthetic Simulatability}}                                                                                                                                                  \\ \hline
\textbf{Dataset}               & \textbf{Model} & \multicolumn{1}{c|}{\textbf{DeepAR}}   & \multicolumn{1}{c|}{\textbf{PatchTST}} & \multicolumn{1}{c|}{\textbf{Moirai}}   & \multicolumn{1}{c|}{\textbf{ETS}}      & \multicolumn{1}{c|}{\textbf{Arima}} & \multicolumn{1}{c|}{\textbf{DeepAR}}   & \multicolumn{1}{c|}{\textbf{PatchTST}} & \multicolumn{1}{c|}{\textbf{Moirai}}   & \multicolumn{1}{c|}{\textbf{ETS}}      & \multicolumn{1}{c|}{\textbf{Arima}} \\ \hline
\multirow{4}{*}{\textbf{Tou.}} & LLMTime        & \multicolumn{1}{r|}{4.55E-01}          & \multicolumn{1}{r|}{4.76E-01}          & \multicolumn{1}{r|}{6.66E-01}          & \multicolumn{1}{r|}{4.16E-01}          & 4.51E-01                            & \multicolumn{1}{r|}{4.87E-01}          & \multicolumn{1}{r|}{5.45E-01}          & \multicolumn{1}{r|}{6.53E-01}          & \multicolumn{1}{r|}{4.16E-01}          & 4.07E-01                            \\
                               & LLMTime\_R     & \multicolumn{1}{r|}{3.41E-01}          & \multicolumn{1}{r|}{3.77E-01}          & \multicolumn{1}{r|}{5.36E-01}          & \multicolumn{1}{r|}{3.62E-01}          & 4.13E-01                            & \multicolumn{1}{r|}{2.87E-01}          & \multicolumn{1}{r|}{3.36E-01}          & \multicolumn{1}{r|}{3.88E-01}          & \multicolumn{1}{r|}{1.62E-01}          & 1.67E-01                            \\
                               & LLMTime\_M     & \multicolumn{1}{r|}{1.44E+00}          & \multicolumn{1}{r|}{1.98E+00}          & \multicolumn{1}{r|}{1.99E+00}          & \multicolumn{1}{r|}{1.99E+00}          & 1.99E+00                            & \multicolumn{1}{r|}{2.00E+00}          & \multicolumn{1}{r|}{2.00E+00}          & \multicolumn{1}{r|}{2.00E+00}          & \multicolumn{1}{r|}{2.00E+00}          & 2.00E+00                            \\
                               & LLMTime\_E     & \multicolumn{1}{r|}{\textbf{2.91E-01}} & \multicolumn{1}{r|}{\textbf{3.35E-01}} & \multicolumn{1}{r|}{\textbf{5.23E-01}} & \multicolumn{1}{r|}{\textbf{2.96E-01}} & \textbf{3.76E-01}                   & \multicolumn{1}{r|}{\textbf{2.05E-01}} & \multicolumn{1}{r|}{\textbf{2.96E-01}} & \multicolumn{1}{r|}{\textbf{3.87E-01}} & \multicolumn{1}{r|}{\textbf{8.80E-02}} & \textbf{7.10E-02}                   \\ \hline
\multirow{4}{*}{\textbf{M3}}   & LLMTime        & \multicolumn{1}{r|}{4.16E-01}          & \multicolumn{1}{r|}{5.10E-01}          & \multicolumn{1}{r|}{7.30E-01}          & \multicolumn{1}{r|}{5.26E-01}          & 4.67E-01                            & \multicolumn{1}{r|}{4.03E-01}          & \multicolumn{1}{r|}{4.44E-01}          & \multicolumn{1}{r|}{5.47E-01}          & \multicolumn{1}{r|}{3.45E-01}          & 3.50E-01                            \\
                               & LLMTime\_R     & \multicolumn{1}{r|}{3.69E-01}          & \multicolumn{1}{r|}{4.29E-01}          & \multicolumn{1}{r|}{5.99E-01}          & \multicolumn{1}{r|}{4.47E-01}          & 4.35E-01                            & \multicolumn{1}{r|}{2.55E-01}          & \multicolumn{1}{r|}{2.41E-01}          & \multicolumn{1}{r|}{3.49E-01}          & \multicolumn{1}{r|}{1.49E-01}          & 1.52E-01                            \\
                               & LLMTime\_M     & \multicolumn{1}{r|}{1.97E+00}          & \multicolumn{1}{r|}{1.98E+00}          & \multicolumn{1}{r|}{1.98E+00}          & \multicolumn{1}{r|}{1.99E+00}          & 1.98E+00                            & \multicolumn{1}{r|}{2.00E+00}          & \multicolumn{1}{r|}{2.00E+00}          & \multicolumn{1}{r|}{2.00E+00}          & \multicolumn{1}{r|}{2.00E+00}          & 2.00E+00                            \\
                               & LLMTime\_E     & \multicolumn{1}{r|}{\textbf{3.28E-01}} & \multicolumn{1}{r|}{\textbf{3.87E-01}} & \multicolumn{1}{r|}{\textbf{5.87E-01}} & \multicolumn{1}{r|}{\textbf{3.99E-01}} & \textbf{3.93E-01}                   & \multicolumn{1}{r|}{\textbf{1.86E-01}} & \multicolumn{1}{r|}{\textbf{2.08E-01}} & \multicolumn{1}{r|}{\textbf{3.17E-01}} & \multicolumn{1}{r|}{\textbf{7.57E-02}} & \textbf{8.81E-02}                   \\ \hline
\multirow{4}{*}{\textbf{M1}}   & LLMTime        & \multicolumn{1}{r|}{3.85E-01}          & \multicolumn{1}{r|}{4.25E-01}          & \multicolumn{1}{r|}{6.65E-01}          & \multicolumn{1}{r|}{3.24E-01}          & 3.46E-01                            & \multicolumn{1}{r|}{4.10E-01}          & \multicolumn{1}{r|}{4.61E-01}          & \multicolumn{1}{r|}{6.87E-01}          & \multicolumn{1}{r|}{3.58E-01}          & 3.81E-01                            \\
                               & LLMTime\_R     & \multicolumn{1}{r|}{3.27E-01}          & \multicolumn{1}{r|}{3.62E-01}          & \multicolumn{1}{r|}{5.76E-01}          & \multicolumn{1}{r|}{2.43E-01}          & 2.74E-01                            & \multicolumn{1}{r|}{1.79E-01}          & \multicolumn{1}{r|}{6.62E-01}          & \multicolumn{1}{r|}{4.61E-01}          & \multicolumn{1}{r|}{1.53E-01}          & 6.01E-01                            \\
                               & LLMTime\_M     & \multicolumn{1}{r|}{1.98E+00}          & \multicolumn{1}{r|}{1.98E+00}          & \multicolumn{1}{r|}{1.98E+00}          & \multicolumn{1}{r|}{2.00E+00}          & 1.99E+00                            & \multicolumn{1}{r|}{2.00E+00}          & \multicolumn{1}{r|}{2.00E+00}          & \multicolumn{1}{r|}{2.00E+00}          & \multicolumn{1}{r|}{2.00E+00}          & 2.00E+00                            \\
                               & LLMTime\_E     & \multicolumn{1}{r|}{\textbf{2.97E-01}} & \multicolumn{1}{r|}{\textbf{3.26E-01}} & \multicolumn{1}{r|}{\textbf{5.58E-01}} & \multicolumn{1}{r|}{\textbf{2.02E-01}} & \textbf{2.52E-01}                   & \multicolumn{1}{r|}{\textbf{1.41E-01}} & \multicolumn{1}{r|}{\textbf{2.10E-01}} & \multicolumn{1}{r|}{\textbf{4.24E-01}} & \multicolumn{1}{r|}{\textbf{9.58E-02}} & \textbf{9.21E-02}                   \\ \hline
\end{tabular}%
\caption{Sanity check experiments for direct simulatability and synthetic simulatability metrics using sMAPE as the distance measure, lower is better. Results are averaged over three runs. Both metrics reliably distinguish high-quality explanations from poor ones.}
\label{tab:sanity}
\end{table*}

\paragraph{Models}
To test if our evaluation metrics are forecasting method-agnostic, we selected diverse models, including statistical methods like auto ARIMA and auto ETS~\cite{garza2022statsforecast}, deep learning models like DeepAR~\cite{Flunkert2017DeepARPF}, and transformer-based models like Moirai~\cite{woo2024unified} and PatchTST~\cite{Nie2022ATS}. These forecasters made predictions on the datasets, and we generated explanations using the baseline method from~\Cref{sec:baseline} with various SoTA LLMs, both open and closed source. The resulting dataset of time series, forecast, and explanation triplets was used to evaluate our metrics and compare LLMs' explanation capabilities.

For both metrics, we first run sanity checks with GPT-4 generated explanations, using three baselines: \texttt{LLMTime}:  a naive baseline that predicts the forecast without any explanation; \texttt{LLMTime\_R}: which uses the pipeline described in~\Cref{sec:baseline}, but explains a random forecast; \texttt{LLMTime\_M}: which prompts LLMTime to predict a constant value for all steps. We compare these baselines against \texttt{LLMTime\_E} which uses the correct forecast in the pipeline from~\Cref{sec:baseline}.

Through sanity checks, we expect \texttt{LLMTime\_E}, using the correct explanation, to predict the forecast better than the other baselines. Afterward, we extend the experiments to other LLMs to compare how explanations from different models improve forecast prediction. Hyperparameter settings for LLMs are detailed in~\Cref{app:hyperparam}.

We use GPT-4 as the backbone for LLMTime in all experiments because our preliminary experiments showed it consistently benefits from useful explanations. Llama-3 performed better without explanations but showed inferior results when explanations were provided (see~\Cref{app:forecaster_llm} for results), making it unsuitable for our study, which relies on the surrogate benefiting from explanations to test performance metrics.
\paragraph{Evaluation Metrics}

Both our performance metrics measure the distance of the prediction to the black-box model forecast and since the time series data we use and generate have diverse scales, we opt for using scale-independent metrics. Specifically we stick to Symmetric Mean Absolute Percentage Error (sMAPE) for evaluation (see~\Cref{app:add_results} for results with more metrics).

\subsection{Results}
\subsubsection{Sanity Checks}
\Cref{tab:sanity} presents the results of the sanity check experiments for both simulatability metrics, averaged over three runs. Notably, pre-pending textual data improves forecasting, regardless of whether the forecast is actual or random (\textit{cf.} LLMTime vs. LLMTime\_R and LLMTime\_E). This likely stems from the engineered nature of our explanation pipeline, which is designed to aid forecasting.

\begin{table*}[!htbp]
\tiny
\centering
\resizebox{\textwidth}{!}{%
\begin{tabular}{|c|l|ccccc|ccccc|}
\hline
\multicolumn{1}{|l|}{}                 &                                                                        & \multicolumn{5}{c|}{\textbf{Direct Simulatability}}                                                                                                                                   & \multicolumn{5}{c|}{\textbf{Synthetic Simulatability}}                                                                                                                     \\ \hline
\multicolumn{1}{|l|}{\textbf{Dataset}} & \textbf{\begin{tabular}[c]{@{}l@{}}Backbone \\ Explainer\end{tabular}} & \multicolumn{1}{c|}{\textbf{DeepAR}}   & \multicolumn{1}{c|}{\textbf{PatchTST}} & \multicolumn{1}{c|}{\textbf{Moirai}}   & \multicolumn{1}{c|}{\textbf{ETS}}      & \textbf{Arima}    & \multicolumn{1}{c|}{\textbf{DeepAR}} & \multicolumn{1}{c|}{\textbf{PatchTST}} & \multicolumn{1}{c|}{\textbf{Moirai}} & \multicolumn{1}{c|}{\textbf{ETS}}  & \textbf{Arima} \\ \hline
\multirow{5}{*}{Tou.}                  & GPT4                                                                   & \multicolumn{1}{c|}{\textbf{1.98E-01}} & \multicolumn{1}{c|}{\textbf{2.93E-01}} & \multicolumn{1}{c|}{\textbf{3.92E-01}} & \multicolumn{1}{c|}{\textbf{9.04E-02}} & \textbf{6.71E-02} & \multicolumn{1}{c|}{\textbf{0.42}}   & \multicolumn{1}{c|}{\textbf{0.43}}     & \multicolumn{1}{c|}{\textbf{0.44}}   & \multicolumn{1}{c|}{\textbf{0.41}} & \textbf{0.45}  \\
                                       & Llama3-70b                                                             & \multicolumn{1}{c|}{{\ul 2.41E-01}}    & \multicolumn{1}{c|}{{\ul 3.31E-01}}    & \multicolumn{1}{c|}{{\ul 4.14E-01}}    & \multicolumn{1}{c|}{{\ul 1.49E-01}}    & {\ul 1.15E-01}    & \multicolumn{1}{c|}{{\ul 0.43}}      & \multicolumn{1}{c|}{{\ul 0.45}}        & \multicolumn{1}{c|}{{\ul 0.44}}      & \multicolumn{1}{c|}{{\ul 0.43}}    & {\ul 0.46}     \\
                                       & Llama2-70b                                                             & \multicolumn{1}{c|}{3.73E-01}          & \multicolumn{1}{c|}{4.63E-01}          & \multicolumn{1}{c|}{5.53E-01}          & \multicolumn{1}{c|}{3.39E-01}          & 3.34E-01          & \multicolumn{1}{c|}{0.49}            & \multicolumn{1}{c|}{0.52}              & \multicolumn{1}{c|}{0.48}            & \multicolumn{1}{c|}{0.50}          & 0.50           \\
                                       & Vicuna-7b                                                              & \multicolumn{1}{c|}{3.15E-01}          & \multicolumn{1}{c|}{3.81E-01}          & \multicolumn{1}{c|}{4.96E-01}          & \multicolumn{1}{c|}{2.15E-01}          & 1.91E-01          & \multicolumn{1}{c|}{0.48}            & \multicolumn{1}{c|}{0.50}              & \multicolumn{1}{c|}{0.50}            & \multicolumn{1}{c|}{{\ul 0.43}}    & 0.48           \\
                                       & Mistral-7b                                                             & \multicolumn{1}{c|}{2.83E-01}          & \multicolumn{1}{c|}{3.88E-01}          & \multicolumn{1}{c|}{4.71E-01}          & \multicolumn{1}{c|}{2.53E-01}          & 2.53E-01          & \multicolumn{1}{c|}{0.44}            & \multicolumn{1}{c|}{0.45}              & \multicolumn{1}{c|}{0.46}            & \multicolumn{1}{c|}{0.44}          & 0.47           \\ \hline
\multirow{5}{*}{M3}                    & GPT4                                                                   & \multicolumn{1}{c|}{\textbf{1.83E-01}} & \multicolumn{1}{c|}{\textbf{2.09E-01}} & \multicolumn{1}{c|}{\textbf{2.88E-01}} & \multicolumn{1}{c|}{\textbf{6.83E-02}} & \textbf{8.54E-02} & \multicolumn{1}{c|}{\textbf{0.42}}   & \multicolumn{1}{c|}{\textbf{0.44}}     & \multicolumn{1}{c|}{\textbf{0.45}}   & \multicolumn{1}{c|}{\textbf{0.42}} & 0.46           \\
                                       & Llama3-70b                                                             & \multicolumn{1}{c|}{{\ul 2.13E-01}}    & \multicolumn{1}{c|}{{\ul 2.34E-01}}    & \multicolumn{1}{c|}{{\ul 2.98E-01}}    & \multicolumn{1}{c|}{{\ul 1.14E-01}}    & {\ul 1.03E-01}    & \multicolumn{1}{c|}{{\ul 0.42}}      & \multicolumn{1}{c|}{{\ul 0.44}}        & \multicolumn{1}{c|}{{\ul 0.46}}      & \multicolumn{1}{c|}{0.45}          & \textbf{0.42}  \\
                                       & Llama2-70b                                                             & \multicolumn{1}{c|}{3.65E-01}          & \multicolumn{1}{c|}{3.61E-01}          & \multicolumn{1}{c|}{5.11E-01}          & \multicolumn{1}{c|}{3.07E-01}          & 3.20E-01          & \multicolumn{1}{c|}{0.52}            & \multicolumn{1}{c|}{0.49}              & \multicolumn{1}{c|}{0.49}            & \multicolumn{1}{c|}{0.50}          & 0.49           \\
                                       & Vicuna-7b                                                              & \multicolumn{1}{c|}{2.96E-01}          & \multicolumn{1}{c|}{2.80E-01}          & \multicolumn{1}{c|}{4.19E-01}          & \multicolumn{1}{c|}{1.76E-01}          & 2.11E-01          & \multicolumn{1}{c|}{0.48}            & \multicolumn{1}{c|}{0.47}              & \multicolumn{1}{c|}{{\ul 0.46}}      & \multicolumn{1}{c|}{{\ul 0.44}}    & {\ul 0.43}     \\
                                       & Mistral-7b                                                             & \multicolumn{1}{c|}{2.21E-01}          & \multicolumn{1}{c|}{2.70E-01}          & \multicolumn{1}{c|}{3.74E-01}          & \multicolumn{1}{c|}{2.45E-01}          & 2.22E-01          & \multicolumn{1}{c|}{0.44}            & \multicolumn{1}{c|}{0.46}              & \multicolumn{1}{c|}{{\ul 0.46}}      & \multicolumn{1}{c|}{0.46}          & 0.51           \\ \hline
\multirow{5}{*}{M1}                    & GPT4                                                                   & \multicolumn{1}{c|}{\textbf{1.47E-01}} & \multicolumn{1}{c|}{\textbf{2.02E-01}} & \multicolumn{1}{c|}{\textbf{4.06E-01}} & \multicolumn{1}{c|}{\textbf{9.80E-02}} & {\ul 1.03E-01}    & \multicolumn{1}{c|}{0.47}            & \multicolumn{1}{c|}{{\ul 0.41}}        & \multicolumn{1}{c|}{\textbf{0.44}}   & \multicolumn{1}{c|}{0.46}          & {\ul 0.44}     \\
                                       & Llama3-70b                                                             & \multicolumn{1}{c|}{1.65E-01}          & \multicolumn{1}{c|}{2.61E-01}          & \multicolumn{1}{c|}{{\ul 4.25E-01}}    & \multicolumn{1}{c|}{{\ul 1.00E-01}}    & \textbf{8.90E-02} & \multicolumn{1}{c|}{{\ul 0.45}}      & \multicolumn{1}{c|}{0.44}              & \multicolumn{1}{c|}{{\ul 0.46}}      & \multicolumn{1}{c|}{\textbf{0.37}} & \textbf{0.40}  \\
                                       & Llama2-70b                                                             & \multicolumn{1}{c|}{4.72E-01}          & \multicolumn{1}{c|}{4.35E-01}          & \multicolumn{1}{c|}{5.93E-01}          & \multicolumn{1}{c|}{3.99E-01}          & 3.80E-01          & \multicolumn{1}{c|}{0.53}            & \multicolumn{1}{c|}{0.50}              & \multicolumn{1}{c|}{0.52}            & \multicolumn{1}{c|}{0.49}          & 0.48           \\
                                       & Vicuna-7b                                                              & \multicolumn{1}{c|}{2.92E-01}          & \multicolumn{1}{c|}{2.94E-01}          & \multicolumn{1}{c|}{5.92E-01}          & \multicolumn{1}{c|}{2.07E-01}          & 2.48E-01          & \multicolumn{1}{c|}{0.48}            & \multicolumn{1}{c|}{0.48}              & \multicolumn{1}{c|}{0.50}            & \multicolumn{1}{c|}{{\ul 0.44}}    & 0.49           \\
                                       & Mistral-7b                                                             & \multicolumn{1}{c|}{{\ul 1.62E-01}}    & \multicolumn{1}{c|}{{\ul 2.46E-01}}    & \multicolumn{1}{c|}{5.04E-01}          & \multicolumn{1}{c|}{2.00E-01}          & 1.66E-01          & \multicolumn{1}{c|}{\textbf{0.39}}   & \multicolumn{1}{c|}{\textbf{0.40}}     & \multicolumn{1}{c|}{0.47}            & \multicolumn{1}{c|}{0.46}          & 0.47           \\ \hline
\end{tabular}%
}
\caption{Performance comparison across different LLMs as baseline explainer backbone using sMAPE as the distance measure, lower is better. The results for synthetic simulatability are normalized by LLMTime performance to ensure a fair comparison across different LLMs. Best results \textbf{bolded}, second best results {\ul underlined}.}
\label{tab:llms}
\end{table*}

As expected, LLMTime\_M, with an adversarial prompt forcing arbitrary predictions, shows greater deviation from the original forecast, impairing results. In contrast, LLMTime\_E, seeded with the correct forecast, consistently yields the closest predictions to the black-box model. This demonstrates that both simulatability metrics effectively distinguish between good and bad explanations.

\begin{figure*}[ht!]
    \centering
    \begin{subfigure}[t]{0.48\linewidth}
        \includegraphics[width=\linewidth]{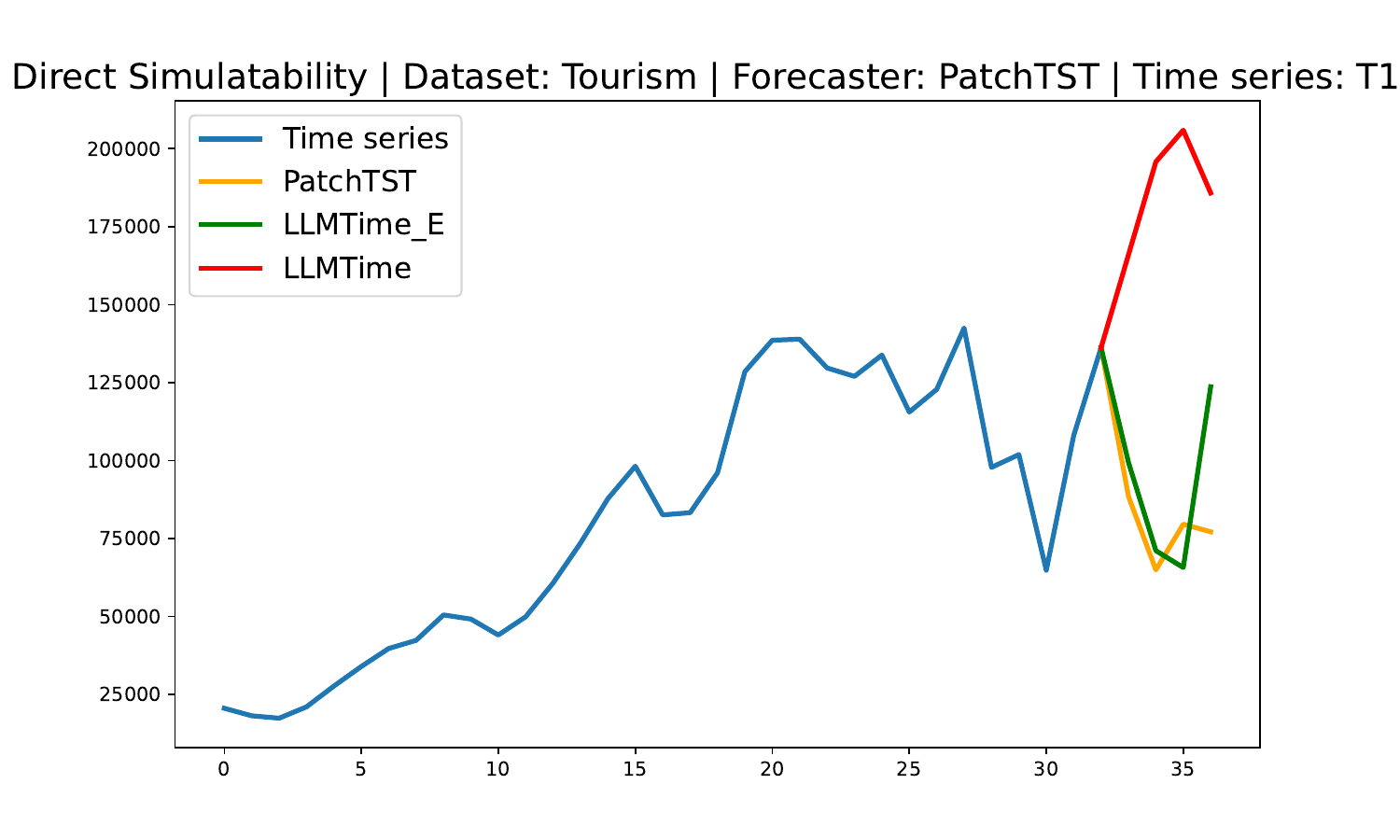}
        \caption{\textbf{Forecast Explanation:} The forecasted data suggests a continuation of the moderate declining trend observed at the end of the provided time series history... the pattern indicates oscillations that may lead to brief recoveries or less pronounced falls in the immediate future steps. Metric results: LLMTime = $0.91$, LLMTime\_E = $0.75$.}
        \label{fig:qual_2}
    \end{subfigure}
    \quad % Space between the first and second subfigures in the first row
    \begin{subfigure}[t]{0.48\linewidth}
        \includegraphics[width=\linewidth]{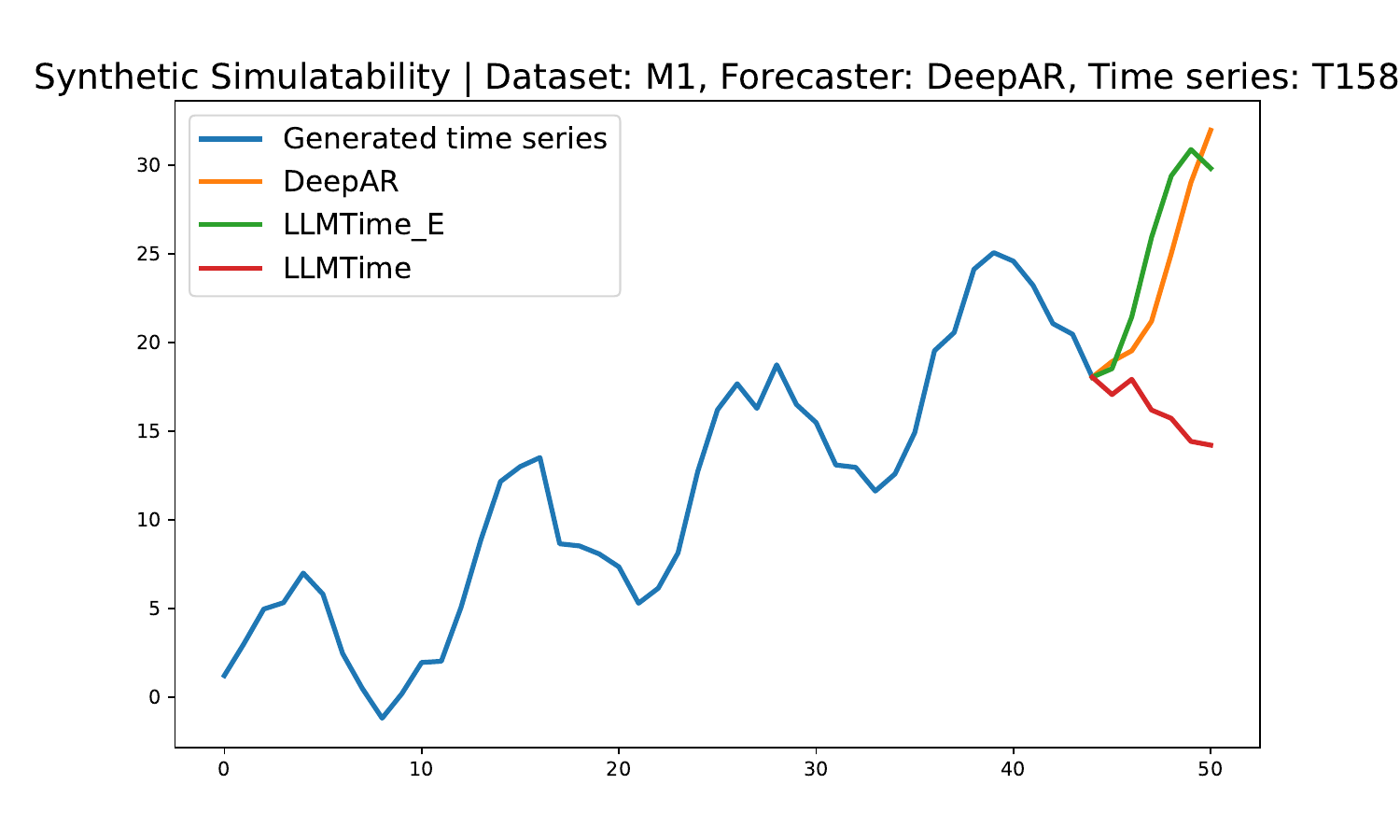}
        \caption{\textbf{Forecast Explanation:} The forecasted data indicates a continuation of the cyclical pattern ... The variability does not show major departures from previous segments, suggesting that the forecaster expects the recent past volatility level to be maintained... Metric results: LLMTime = $0.25$, LLMTime\_E = $0.22$.} 
        \label{fig:qual_3}
    \end{subfigure}
    \caption{Qualitative examples showing effect of explanation on the forecast. 
    \Cref{fig:qual_2,fig:qual_3} show examples for direct and synthetic simulatability respectively.
    Historical context shown in \textcolor{blue}{blue}, black-box model forecast shown in \textcolor{orange}{orange}, LLMTime prediction shown in \textcolor{red}{red}, and LLMTime\_E prediction shown in \textcolor{green}{green}.
    }
    \label{fig:qual_main}
\end{figure*}

\subsubsection{Qualitative Examples}
In this section, we examine two qualitative examples in~\Cref{fig:qual_main}, for direct and synthetic simulatability. Since these metrics quantify how helpful the explanations are in predicting a model's forecast, it is crucial to understand their meaningful impact on predictions. We use diverse datasets and forecasters in these examples to demonstrate the evaluation metrics' capabilities across different setups. More examples can be found in Appendix--~\Cref{fig:qual_extra}.

\paragraph{Direct Simulatability} \Cref{fig:qual_2}  shows an example from the Tourism dataset using the PatchTST model. LLMTime
forecast assumes the continuation of the recent rising trend. However, with the explanation, which suggests a declining trend with patterns of oscillations, LLMTime\_E's prediction aligns better with the behavior of PatchTST.

\paragraph{Synthetic Simulatability}  For synthetic simulatability, the time series is generated at runtime based on the explanation. There are two aspects to analyze: $(i)$ whether the generated time series accurately represents the explanation, and $(ii)$ whether the explanation improves the prediction of the forecaster's output. We observe that the generated time series in~\Cref{fig:qual_3} contains cyclical patterns as mentioned in the explanation, and the ground truth forecast maintains these cycles as suggested. Also LLMTime\_E's generation, which utilizes the explanation, aligns more closely with the DeepAR forecast.

Overall, we observe that high-quality explanations help the language model make better predictions of the black-box forecasting model's output.

\subsubsection{Comparison Across Different LLM Explainers}

Next, we compare different backbone LLMs used with our baseline explainer. For synthetic simulatability, since time series are generated at runtime, comparing sMAPE across explainers is unfair due to sample variability. To ensure fairness, we normalize results by LLMTime performance:
\begin{equation}
\tiny
NSS(LLMTime_E) = \frac{SS(LLMTime_E)}{SS(LLMTime) + SS(LLMTime_E)}
\end{equation}
where $SS$ and $NSS$ stands for synthetic and normalized synthetic simulatability respectively.

We also emphasize that direct and synthetic simulatability are not directly comparable, as they evaluate performance on different time series. While synthetic simulatability may seem harder due to loosely related time series, the generated task could still be simpler. The key comparison in this section focuses on which LLM improves the forecast the most in both direct and synthetic simulatability, evaluated separately.

~\Cref{tab:llms}  presents the results, GPT-4 produces the best explanations for predicting black-box model forecasts, with Llama3-70b performing best among open-source models. According to both metrics, GPT-4 explanations most effectively help predict black-box model forecasts in most cases. However, since LLMTime also uses GPT-4 as the backbone model, these results might be influenced by alignment (refer to ~\Cref{app:forecaster_llm} on why we stick to GPT-4 for all experiments). The performance difference between Llama2 and Llama3 aligns with findings that the latter excels in math reasoning benchmarks like GSM8K and MATH~\cite{open-llm-leaderboard}. Although not directly transferable, we believe high reasoning performance on numerical tasks indicates better reasoning on time series data.

\paragraph{Model Size and Performance.} Judging by the Llama2-70B results, model size alone does not guarantee high-quality explanations. For example, Vicuna-7b outperforms Llama2-70B on most dataset-forecaster pairs across both metrics, despite having ten times fewer parameters. This suggests that numerical reasoning may correlate with better time series reasoning, as Vicuna-7b has demonstrated stronger numerical reasoning compared to larger models~\cite{Zheng2023JudgingLW}.

\paragraph{Explanations Across Forecaster Families:}
A key question is whether explanations behave similarly across model families. We use the direct simulatability metric for comparison, as it uses the same time series to simulate each forecaster’s predictions. As shown in~\Cref{tab:llms}, PatchTST and Moirai have the largest error values for simulatability, likely due to patch embeddings and our use of short time series sequences. When the context is shorter than the minimum patch size, these models underperform, affecting explanation quality. In contrast, statistical models show the smallest error, likely because their simpler behavior is easier to explain in natural language.

\subsubsection{Human Study}
We conduct a two-part human study to evaluate the effectiveness of our proposed metrics in distinguishing useful explanations, see~\Cref{app:human_study_details} for setup details. We sample 20 examples where both direct and synthetic simulatability metrics agree on whether an explanation is "useful" or "not useful." An explanation is deemed useful if it improves time series forecast accuracy, and these examples are used in both parts of the study.

In the first part, participants replaced the human surrogate in~\Cref{fig:metrics}, and were asked to draw forecast for 20 time series, initially without any explanation, and then with explanations. We calculated the average improvement in the forecast for both useful and non-useful explanations. The results show that explanations deemed useful by our metrics improved accuracy by 5\%, while non-useful explanations reduced it by -11\%. This validates our metrics' ability to identify helpful explanations, with a Kappa score of 0.42 showing moderate agreement between user-improving explanations and those classified as useful.

In the second part of the study, participants were shown the ground truth forecast and asked to assess the usefulness of the explanation. This evaluation yielded a Kappa score of 0.58, indicating moderate agreement with our metrics. These results confirm that human judgment aligns well with our metrics, validating their reliability in distinguishing between useful and non-useful explanations.

\section{Future Directions}
\paragraph{Time-series and language foundation models} This study is a first step in bridging time-series and language foundation models. Future work could explore using NLEs to fine-tune models capable of processing both modalities.

\paragraph{Applications of performance metrics} Our metrics can be applied beyond evaluation, such as in n-best sampling or as rewards for fine-tuning LLMs to generate better explanations, similar to ~\citet{Chen2024TowardsCN}'s work.

\paragraph{Other downstream time-series tasks} While we focus on forecasting, future work could explore NLE generation and evaluation for other tasks like time-series classification and anomaly detection.

\section{Conclusion}
We propose two complementary metrics for evaluating time series forecasting explanations: $i)$ direct simulatability, which measures how well explanations help predict the black-box model’s forecast on the original time series, and $ii)$ synthetic simulatability, which evaluates how well the explanation generalizes to synthetic time series generated from the explanation. Through sanity check experiments, we demonstrate that both metrics can distinguish between good and poor explanations. We then use these metrics to evaluate various LLMs' explanation performance. Our findings are two-fold: first, model size impacts explanation performance, but numerical reasoning capability is more critical; second, explanations for statistical model forecasters provide more insight than those for deep learning and transformer-based forecasters. A human study shows high agreement between our metrics and human annotators, supporting our premise that explanations must help humans understand the model's behavior. Finally, qualitative analysis shows that high-quality explanations improve LLM predictions, validating our metrics' effectiveness. We hope this study encourages further exploration of natural language explanation pipelines for time series forecasting.
\newpage
\clearpage  

\section{Limitations}
The use of LLMs in time series forecasting has gained popularity only recently and still faces many limitations. Moreover generating natural language explanations poses challenges in itself, irrespective of the target modality. As our proposed project is the first to explore the intersection of these two fields, there are numerous areas for potential improvement. This section discusses these areas of limitations:

\noindent\textbf{1) Lack of Multivariate TS.} Due to its challenging nature, this work omits experiments for multivariate time series forecasts. Future research could explore explanations for multivariate time series forecasts, which would involve feeding the LLM with multiple series and requiring the explainer to perform cross-variate reasoning.

\noindent\textbf{2) Limited Frequency and Context Lengths.} Due to the limitations in processing long series in LLMs, our experiments are restricted to short context and low-frequency time series data. As LLMs improve in reasoning capabilities and handling longer numerical contexts, our methods for both explanation and evaluation can be applied to lower frequencies.

% Already discussed in future work
% \noindent\textbf{3) Applications to Multi-modal Research.} This work does not explore the application of paired data we generated for other tasks, such as judgmental forecasting, which uses external textual information for making predictions. Future research could investigate this potential avenue.

% Entries for the entire Anthology, followed by custom entries
\bibliography{anthology,custom}

\appendix

\section{Choice of Forecaster LLM}
\label{app:forecaster_llm}
\begin{table}[htb!]
\resizebox{0.48\textwidth}{!}{%
\begin{tabular}{|c|c|c|c|c|}
\hline
\textbf{Forecaster LLM}           & \textbf{Model} & \textbf{DeepAR}   & \textbf{PatchTST} & \textbf{Arima}    \\ \hline
\multirow{2}{*}{\textbf{Llama-3}} & LLMTime        & 1.90E-01          & \textbf{2.58E-01} & 1.70E-01          \\ \cline{2-5} 
                                  & LLMTime\_E     & \textbf{1.83E-01} & 2.71E-01          & \textbf{1.60E-01} \\ \hline
\multirow{2}{*}{\textbf{GPT-4}}   & LLMTime        & 4.67E-01          & 4.53E-01          & 4.03E-01          \\ \cline{2-5} 
                                  & LLMTime\_E     & \textbf{1.37E-01} & \textbf{2.07E-01} & \textbf{8.67E-02} \\ \hline
\end{tabular}%
}
\caption{Performance comparison for Llama-3-70b and GPT-4 as the forecaster LLM on M1 dataset.}
\label{tab_app:forecaster_llm}
\end{table}

Before scaling up our experiments to all datasets, we conducted a small experiment to compare the performance of Llama-3 and GPT-4 for forecasting on the M1 dataset. According to~\Cref{tab_app:forecaster_llm} although Llama-3 outperformed GPT-4 in forecasting without explanations, its performance with useful explanations was either marginal or inferior. Specifically, it struggled to leverage explanations effectively, yielding only minor improvements or, in some cases, worse results compared to GPT-4. Since the primary aim of our study is to test performance metrics that depend on a strong assumption that the surrogate can benefit from explanations, we found Llama-3 unsuitable for this role. In contrast, GPT-4 consistently showed improvements when provided with useful explanations, aligning with the goal of our simulatability metrics. For this reason, we opted for GPT-4, despite it being a closed-source model. It is important to note that our pipeline is flexible and can be adapted to work with any open-source model as advancements in the field lead to improvements in time series reasoning.
% Omitting this for now.
% \section{The effect of adding text to LLMTime Prompt}
% \label{app:LLMtime_w_text}
% \ta{Add qualitative examples showing how the addition of text play a stabilizer role for LLMTime}.
\section{Evaluation Prompts}
\label{app:eval_prompts}
This section lists the specific prompts that we use for evaluation pipeline depicted in~\Cref{sec:metrics}.

\paragraph{Direct Simulatability}
For direct simulatability we use LLM as human surrogate to predict the forecast predictions both with and without explanation to generate results respectively for LLMTime and LLMTime\_E (\textit{cf.}~\Cref{subsec:exp_setup}). The prompt for generating the forecast without explanation largely follows the suggested prompt by~\cite{gruver2023llmtime} and is shown in~\Cref{tab:LLMTIME_prompt}. The prompt for generating the forecast grounded on the explanation is shown in~\Cref{tab:LLMTIME_exp_prompt}. The main difference between this prompt and the former is the inclusion of a forecast tip, which is generated from the forecast explanation using the prompt in~\Cref{tab:forecast_tip_prompt}. 
\begin{table}
    \centering
    \begin{tabular}{p{7cm}}
    
    \cellcolor{yellow!10}You are a helpful assistant that performs time series predictions. The user provide you with a sequence and you will continue the given sequence for \{forecast\_horizon\} steps. The sequence is represented by decimal strings separated by commas. Please continue the sequence without producing any additional text. Do not say anything like 'the next terms in the sequence are', just return the numbers. 
    
    Sequence:
    
    \{time\_series\_data\}
    \end{tabular}
    \caption{The prompt used to generate the forecast prediction without explanation. We follow a similar prompt to LLMTime~\citet{gruver2023llmtime}}
    \label{tab:LLMTIME_prompt}
\end{table}

\begin{table}
    \centering
    \begin{tabular}{p{7cm}}
    
    \cellcolor{yellow!10}You are a helpful assistant that performs time series predictions. The user will provide you with some tips to follow for forecasting and also a sequence. Then you will continue the given sequence for \{forecast\_horizon\} steps. The sequence is represented by decimal strings separated by commas.

    Forecast Tip:
    
    \{forecast\_tip\}

    Please continue the sequence according to the given tips without producing any additional text. Do not say anything like 'the next terms in the sequence are', just return the numbers. 
    Sequence:
    
    \{time\_series\_data\}
    
    \end{tabular}
    \caption{The prompt used to generate the forecast prediction with explanation.}
    \label{tab:LLMTIME_exp_prompt}
\end{table}

\begin{table}
    \centering
    \begin{tabular}{p{7cm}}
    
    \cellcolor{yellow!10}You are given a paragraph that explains the reasoning behind forecasting result of some time series.
    Can you change it such that it is a recommendation to a another user who needs to do forecast on the same time series. Try to keep the recommendations short up to two or three sentences. 
    
    Here is the paragraph:
    
    \{forecast\_explanation\}
    
    \end{tabular}
    \caption{The prompt used to generate the forecast tip from the forecast explanation.}
    \label{tab:forecast_tip_prompt}
\end{table}

\paragraph{Synthetic Simulatability}

\begin{figure*}
    \centering
    \includegraphics[width=0.8\textwidth]{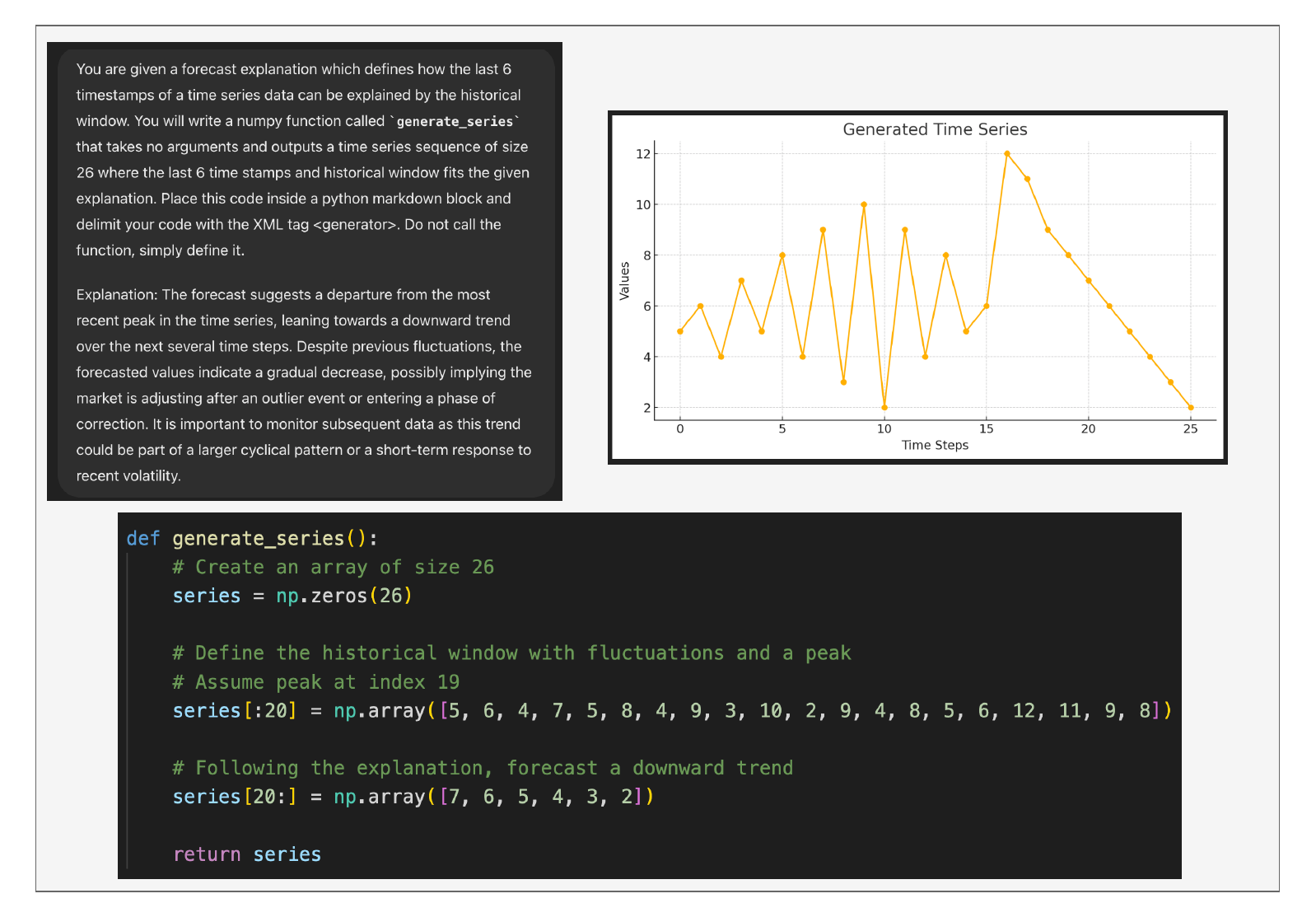}
    \caption{Sample time series generation through code as an intermediary using GPT-4. Using code as an intermediary gives the LLM higher control, allowing it to make changes that directly reflect the explanation.}
    \label{fig:generated_ts}
\end{figure*}

For synthetic simulatability we use the same prompt as direct simulatability to do forecast on the prediction. However, the time series on which the simulation is applied needs to be generated anew using GPT-4. The prompt for such generation is shown in~\Cref{tab:ts_generation_prompt} (mostly inspired by~\citet{Merrill2024LanguageMS}) and a sample with explanation, generated Python code and the resulting time series is shown in~\Cref{fig:generated_ts}.

\begin{table}
    \centering
    \begin{tabular}{p{7cm}}
    
    \cellcolor{yellow!10}You are given a forecast explanation which defines how the last \{forecast\_horizon\} timestamps of a time series data can be explained by the historical window. You will write a numpy function called `generate\_series` that takes no arguments and outputs a time series sequence of size \{timeseries\_size\} where the last \{forecast\_horizon\} time stamps and historical window fits the given explanation. Place this code inside a python markdown block and delimit your code with the XML tag <generator>. Do not call the function, simply define it.

    Explanation: \{forecast\_explanation\}
    \end{tabular}
    \caption{The prompt used to generate the forecast tip from the forecast explanation.}
    \label{tab:ts_generation_prompt}
\end{table}

\section{Explanation Generation Pipeline and Prompts}
\label{app:baseline_details}

In this section, we provide a detailed view of the tools and prompts used to design our baseline in~\Cref{sec:baseline}. The first step in our explanation generation process is segmentation, following the method from \citet{Sharma2021T3DN}, which separates the time series by fitting slopes to each segment. After segmentation, we calculate statistical data for each segment and summarize each segment using a templated string. We then concatenate all segment summaries and prompt the language model to generate a summary for all segments, referred to as $\{segment\_analysis\}$, \textit{cf.} \Cref{tab:segment_prompt}. Next, we provide the LLM with the full time series data along with the $\{segment\_analysis\}$ generated in the previous step and ask it to output a full analysis of the historical context (\textit{cf.}  \Cref{tab:history_exp_prompt}. Finally, using the prompt depicted in \Cref{tab:forecast_exp_prompt} we provide the LLM with the time series data, the summary analysis from the earlier step, the forecast from the black-box forecaster, and a preliminary analysis of the forecast using \cite{ShapTimeZhang}, which highlights the most important indices used for forecasting. The output of this final prompt is our time series forecasting explanation. For a sample of such an explanation, see~\Cref{fig:exp_sample}.
\begin{table}
    \centering
    \begin{tabular}{p{7cm}}
    
    \cellcolor{gray!10}You are a helpful assistant who is expert in understanding time series data.

    You were given some time series data and used an external tool to find different segments in the data along with their slopes and mean and std values.
    Try to understand the segments and their characteristics and generate a brief analysis of the time series' segments such as seasonality, cycles and overall trend.

    Here is an individual analysis of each segment:

    There are \{N\} segments in the time series
    
    1. Segment \{k\} starts at index \{start\_idx\}, ends at index \{end\_idx\}. The mean is \{mean\} the std is \{std\} and the slope in this segment is \{trend\}. It repeats itself every \{seasonality\} predictions.
    
    2...
    
    Based on the above information, generate an analysis with few sentences of the time series' segments with information such as seasonality, cycles and overall trend.
    \end{tabular}
    \caption{The prompt used to aggregate templated segment explanations into final segment analysis.}
    \label{tab:segment_prompt}
\end{table}

\begin{table}
    \centering
    \begin{tabular}{p{7cm}}
    
    \cellcolor{gray!10}You are a helpful assistant who is expert in understanding time series data. You are provided with the full length of time series data in comma separated format, along with some finegranular analysis of the time series data. Then you are asked to generate a short analysis report of the time series data. The report should help the user to forecast the next steps in time series data. Keep the analysis short and to the point. Avoid being redundant and don't suggest that further analysis is needed.
    
    Here is the time series data in comma separated format:
    
    \{time\_series\_data\}
    
    Here is the analysis for all segments:
    
    \{segment\_analysis\}
    
    Generate a short analysis with 2-3 sentences that explain the data in general terms and give hints for the forecaster.
    \end{tabular}
    \caption{The prompt used to generate the final summary of the historical time series data.}
    \label{tab:history_exp_prompt}
\end{table}

\begin{table}
    \centering
    \begin{tabular}{p{7cm}}
    
    \cellcolor{gray!10}You are a helpful assistant who is expert in explaining forecasts of time series data. You are provided with the full length of time series data in comma separated format, along with some finegranular analysis of the time series data. 
    You are also provided with the estimated forecast of the data for the next \{forecast\_horizon\} time steps. 
    
    Then you are asked to generate a short 2-3 sentences long interpretation of the forecasted data. Explain the forecasted data in terms of the time series' temporal structure, variability, long-term memory, trend, and seasonal pattern. Do not use specific data points or numbers in your analysis. Instead, focus on the general trends and patterns in the data.

    Here is the time series in comma separated format:
    
    \{time\_series\_data\}
    
    Here is the analysis of the data:
    
    \{time\_series\_analysis\}
    
    Here is the forecasted data for the next \{forecast\_horizon\} time steps:
    
    \{forecasted\_data\}

    Here is a preanalysis of the forecast:
    
    \{forecast\_preanalysis\}
    
    Generate a short analysis reporting interpretation of the forecasted data with 2-3 sentences.
    \end{tabular}
    \caption{The prompt used to generate the final explanation depicting the interaction between the historical time series data and the forecast.}
    \label{tab:forecast_exp_prompt}
\end{table}

\section{Synthetically Generated Examples}
\label{app:synth_examples}

\begin{figure*}[htbp]
    \centering
    \begin{subfigure}[t]{0.48\textwidth}
        \centering
        \includegraphics[width=\textwidth]{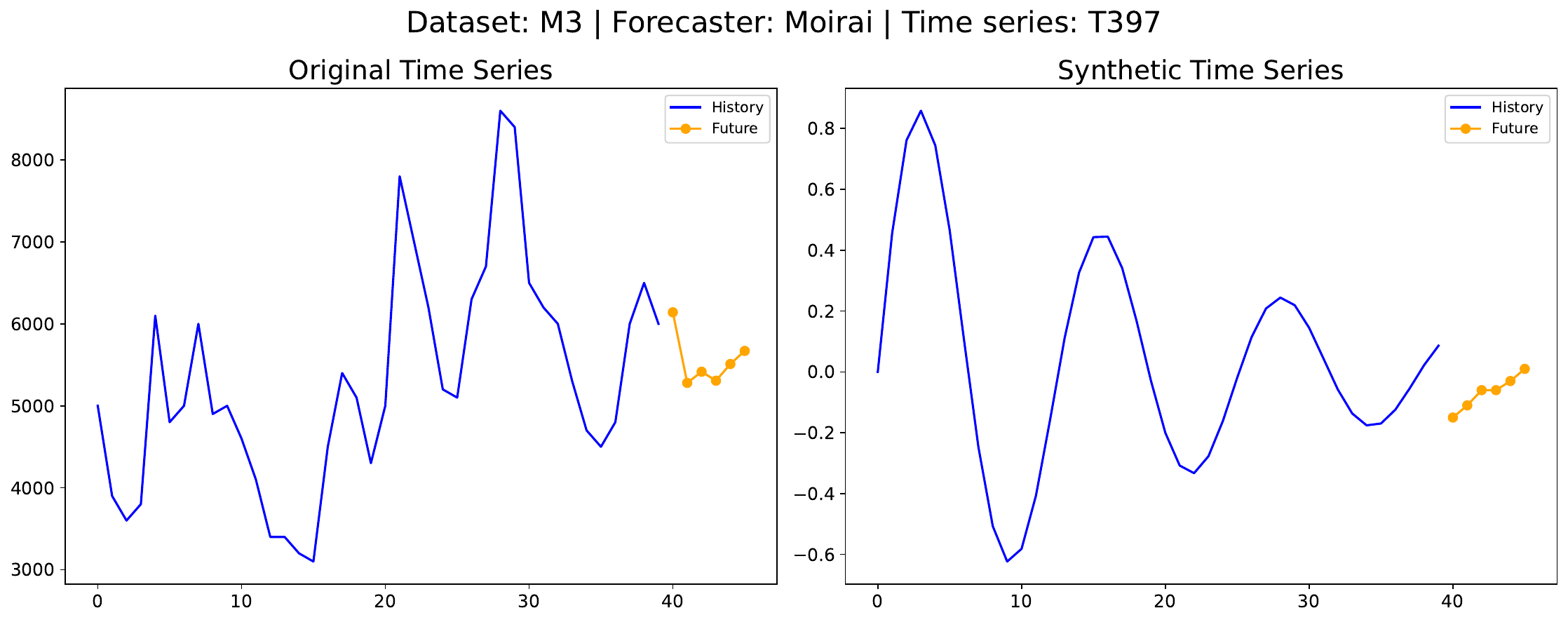}
        \caption{Forecast Explanation:
The forecast suggests that the time series may continue in its cyclical pattern but with a general decline in the amplitude of the cycle, considering the series' progression from a high point toward the latest data provided. The variability seen in the forecasted data indicates some continuation of fluctuation within a narrower band, potentially reflecting less volatility than in the earlier growth phases of the series. While there seems to be no clear long-term trend indicated in the short forecast horizon, a slight upward adjustment toward the end may hint at the beginning of a new growth phase in accordance with the cyclical nature of the series.}
        \label{fig:synth_ex1}
    \end{subfigure}
    \hfill
    \begin{subfigure}[t]{0.48\textwidth}
        \centering
        \includegraphics[width=\textwidth]{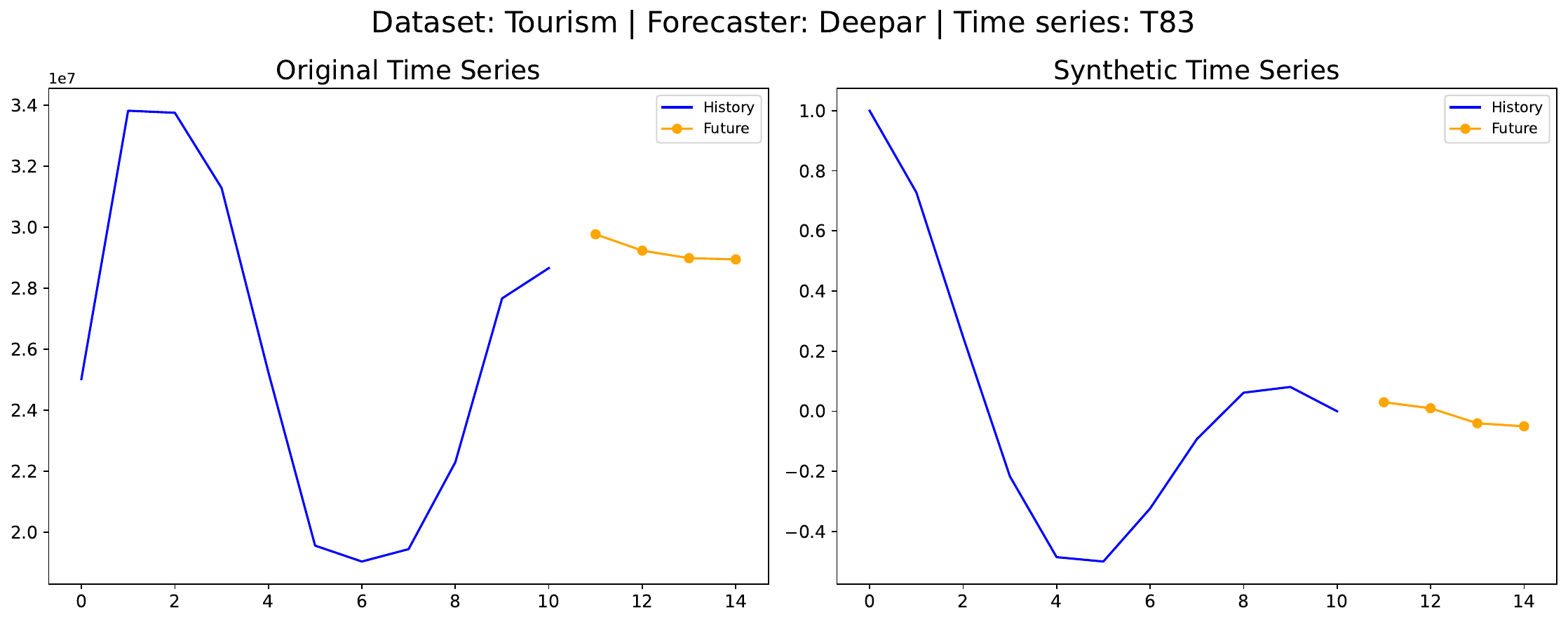}
        \caption{Forecast Explanation: The forecasted data indicates a continuation of the previously identified cyclical pattern, with the series expected to traverse the upward phase of its cycle. Despite this rise, the forecast suggests a gradual leveling off, with the increment between predicted values decreasing over time. This pattern implies that while the series is continuing its recovery from the previous trough, the momentum may be slowing, possibly approaching the peak of the cycle or a plateau.
}
        \label{fig:synth_ex2}
    \end{subfigure}

    \vspace{0.5cm} % Space between rows

    \begin{subfigure}[t]{0.48\textwidth}
        \centering
        \includegraphics[width=\textwidth]{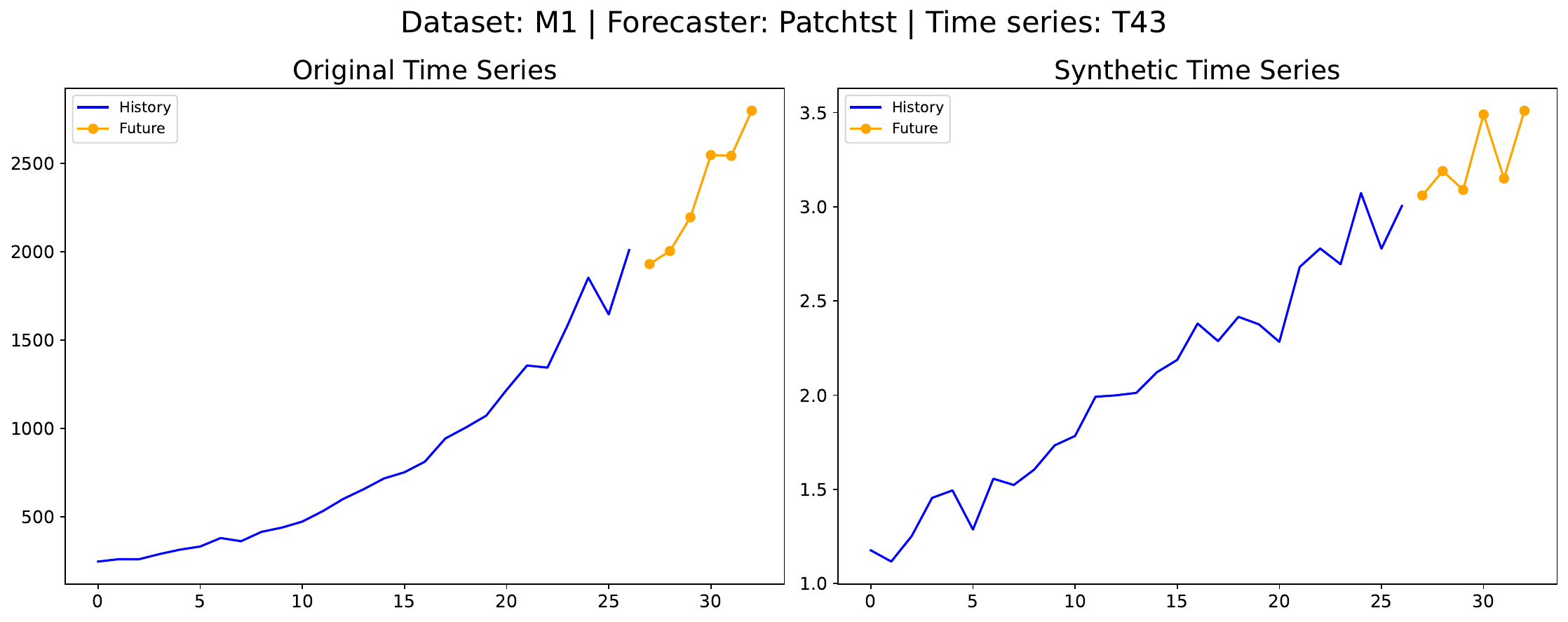}
        \caption{Forecast Explanation:
The forecast predicts a continued upward trend in the time series data, aligned with the strengthening trend observed in the latter part of the provided data. Although the forecast suggests a rise in values, there's potential for some variability, reflecting the fluctuating volatility previously noted. Overall, the forecast points towards an ongoing increase with periods of acceleration, consistent with the time series' recent behaviour.}
        \label{fig:synth_ex3}
    \end{subfigure}
    \hfill
    \begin{subfigure}[t]{0.48\textwidth}
        \centering
        \includegraphics[width=\textwidth]{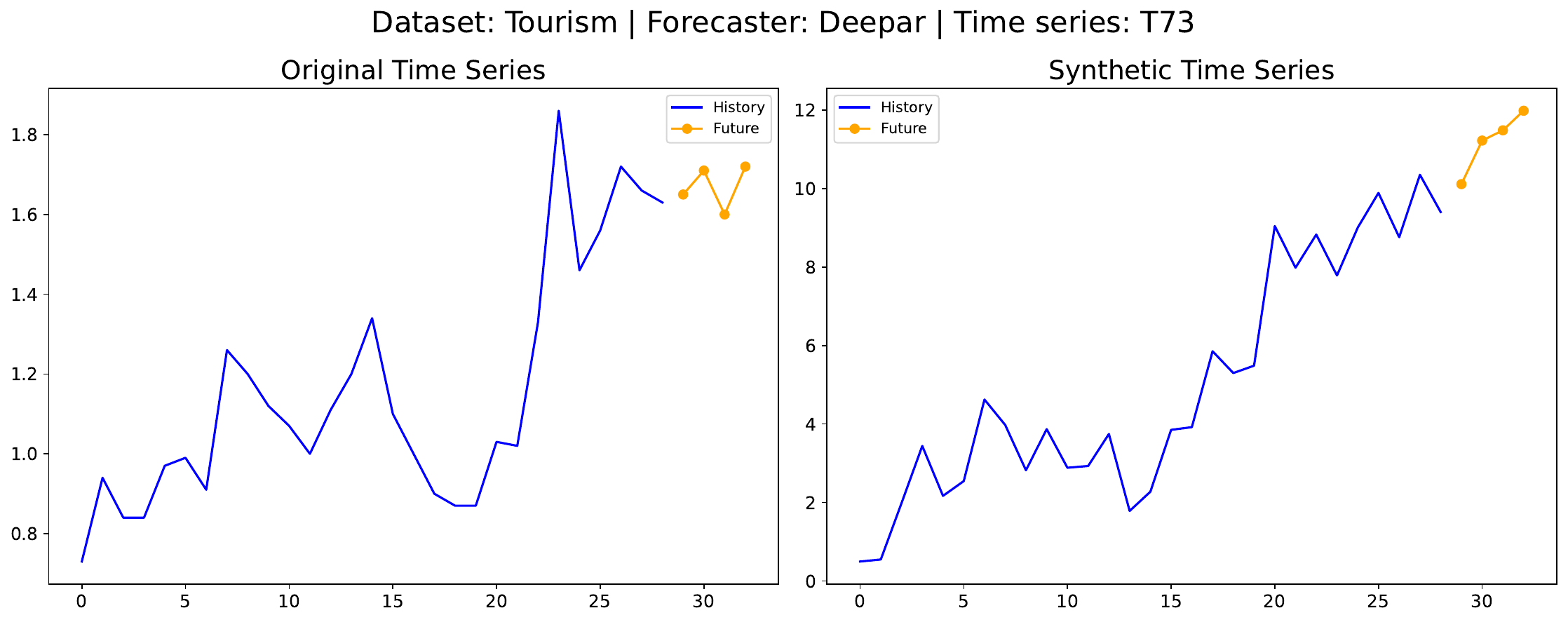}
        \caption{Forecast Explanation:
The forecasted data suggests a continuation of the increasing trend observed in the time series, with some fluctuations that reflect the variability and volatility noted in the most recent segment. The alternating values in the forecast indicate ongoing oscillations around an upward trajectory, maintaining the pattern of growth and contraction. The forecast does not suggest a clear break from the established trend, indicating the series' long-term memory is influencing the near-term outlook.}
        \label{fig:synth_ex4}
    \end{subfigure}

    \caption{Synthetically generated time series alongside their original counterparts and corresponding explanations across different datasets and forecasting models.}
    \label{fig:synth_exs}
\end{figure*}

The synthetic generation process ensures diversity by requiring the large language model (LLM) to create new time series that adhere to the same explanation, without revealing any information about the original series. This approach forces the LLM to generate time series that capture the high-level properties described in the explanation—such as trends, seasonality, or other structural patterns—while differing in aspects like scale, noise, or variability. By doing so, we can assess whether the explanation generalizes broadly to describe the forecaster's behavior or if it is overfitted to the specific values of the original time series. As shown in ~\Cref{fig:synth_exs}, the original and synthetically generated series may differ in their specific data points, but they still follow the same reasoning laid out by the explanation. This diversity is crucial for testing the robustness of the explanation across different yet structurally similar time series, ensuring that it reflects the model’s broader decision-making patterns rather than being overly tailored to one instance.

\begin{figure*}[htb!]
    \centering
    \begin{subfigure}[t]{0.48\linewidth}
        \includegraphics[width=\linewidth]{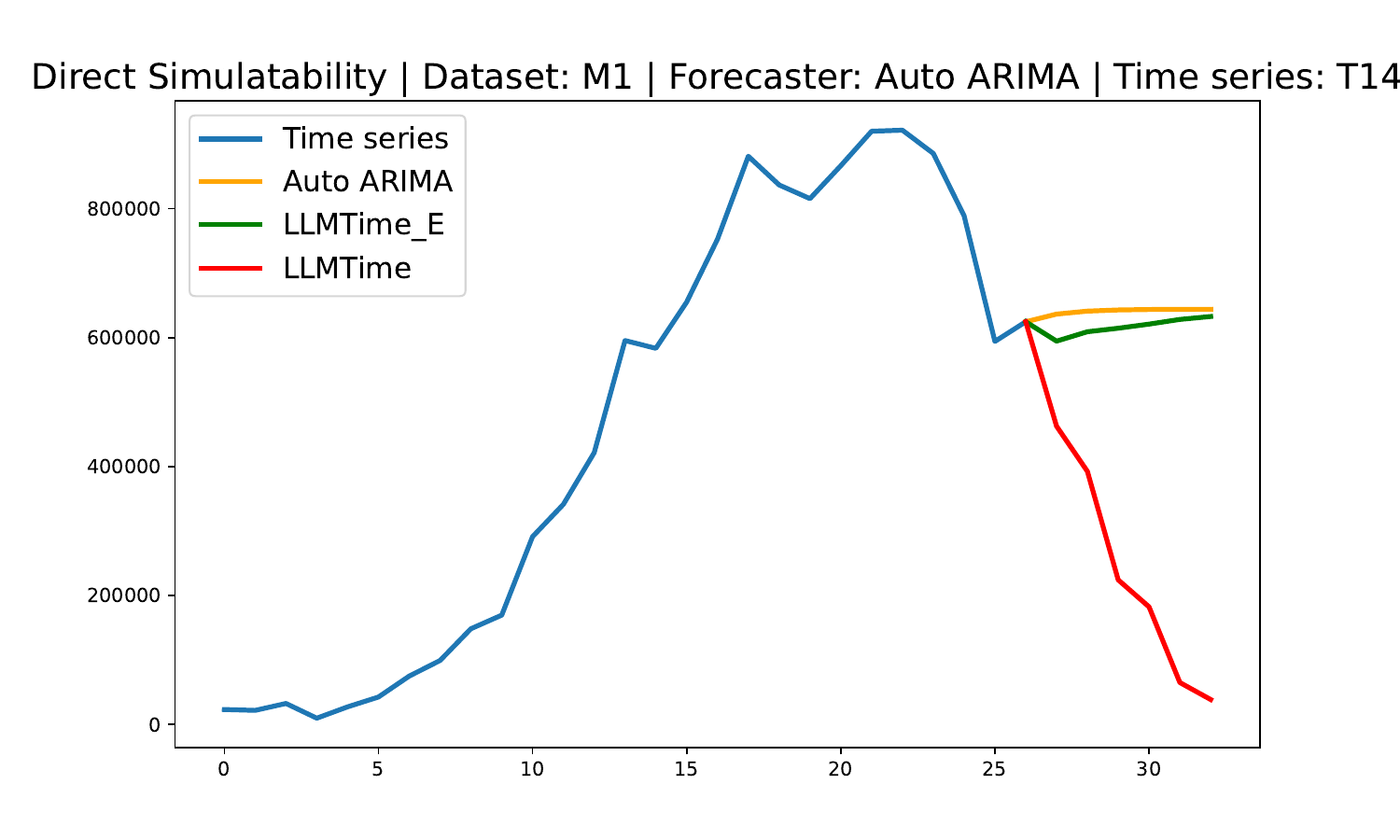}
        \caption{\textbf{Forecast Explanation:} The forecasted data suggests a stabilization in the time series following the recent sharp decline... the rather flat trajectory in the forecast points to a period of low variability, providing no strong signs of an immediate return to the previous high growth rates. Metric results: LLMTime = $1.04$, LLMTime\_E = $0.04$.} 
        \label{fig:qual_1}
    \end{subfigure}
    \quad % Space between the first and second subfigures in the first row
    \begin{subfigure}[t]{0.48\linewidth}
        \includegraphics[width=\linewidth]{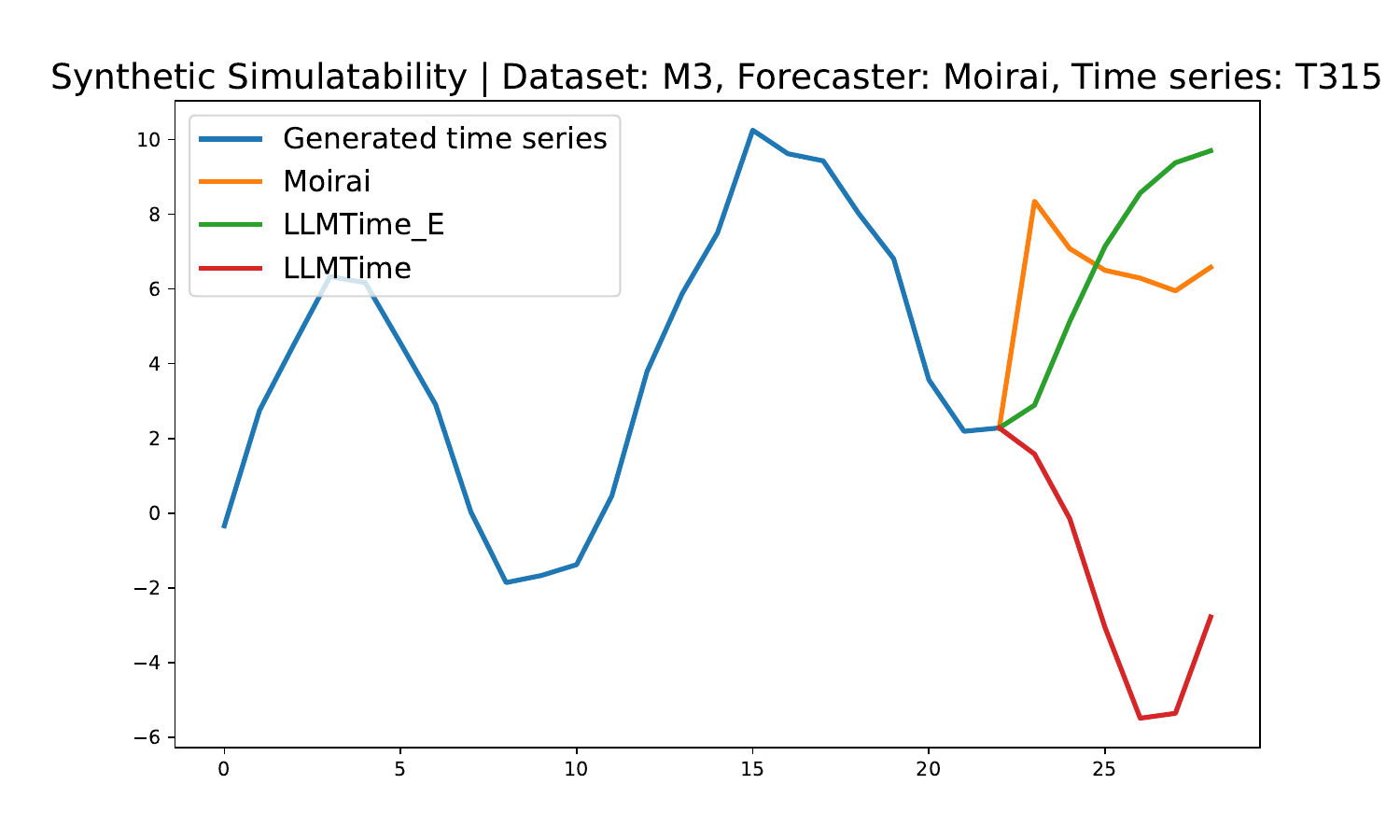}
        \caption{\textbf{Forecast Explanation:} The forecast suggests the time series will continue to display its cyclical nature with variability evident as it includes both rises and dips, indicating ongoing fluctuations within an overall growing trend... Metric results: LLMTime = $1.89$, LLMTime\_E = $0.42$.
}
        \label{fig:qual_4}
    \end{subfigure}

    \caption{More qualitative examples with explanations and their respective effect on forecasting. \Cref{fig:qual_1} shows an example extracted from simulatability experiments, whereas \Cref{fig:qual_4} is from synthetic simulatability experiments. }
    \label{fig:qual_extra}
\end{figure*}
% \section{Leaking Explanation Examples}
% \label{app:leak_examples}

% \begin{table}
%     \centering
%     \begin{tabular}{p{7cm}}
    
%     \cellcolor{blue!10}\textbf{Explanation:} Based on the provided time series data and analysis, \textbf{the forecasted values of 36.85, 39.37, 41.50, 43.85, 46.73, and 48.52 indicate a continuation of the upward trend observed in the time series}. The forecast suggests that the values are likely to increase further in the short term. The variability in the time series data, which was noted as irregular fluctuations, may still be present in the forecasted data, potentially leading to short-term deviations from the overall trend.
%     \end{tabular}
%     \caption{An example of an explanation with leaked information from the forecast. Dataset:M1, Instance ID:T126, Backbone LLM:Mistral}
%     \label{tab:leaking_example}
% \end{table}
% \Cref{tab:leaking_example} shows a sample explanation with forecast leaked, generated using the pipeline depicted in \Cref{sec:baseline} with the Mistral-7b model as the LLM backbone. The leakage problem for direct simulatability has been reiterated in earlier work and undermines the evaluation based solely on this metric. Therefore, we propose synthetic simulatability, which complements the evaluation by assessing simulation performance on different timeseries whose main features align with the given explanation.

% Omitted for now
% \section{Human Study Details}
% \label{app:human_study}

\section{Additional Qualitative Examples}
\label{app:qual_ex_cont}
\Cref{fig:qual_1}  shows an example explanation for a time series in the M1 dataset using the Auto ARIMA forecaster. The naive LLMTime forecast, without an explanation, naturally assumes a continuation of the decreasing trend. However, ARIMA models assume that the data is stationary, leading to a prediction of a flattened line. The figure clearly shows that LLMTime with the explanation does a much better job of predicting the main forecast. It is important to note that, as in earlier XAI work, we are not focused on the accuracy of the prediction itself but on whether the explanation is faithful to the main forecaster.

For \Cref{fig:qual_4} the generated time series depicts a cyclical plot with an overall growing trend as mentioned in the explanation. Regarding the forecasting, it is possible to see that LLMTime forecasting with the explanation again aligns better with the actual forecast compared to the no-explanation generation, which assumes a downward trend following the past few time steps.

\section{Experimental Setup Details and Additional Results}
\label{app:setup_and_results}

\subsection{Hyper parameter setting for LLMs}
\label{app:hyperparam}
We use five different LLMs in our experiments, with the only closed-source model being \texttt{GPT-4} (specifically the `gpt-4-1106-preview' model available via API). For \texttt{GPT-4}, we set the \textit{temperature} to 1.0 following \citet{gruver2023llmtime}. For all open-source LLMs (\texttt{Llama3}, \texttt{Llama2}, \texttt{Mistral}, and \texttt{Vicuna}), we use the following settings: \textit{temperature} of 0.9, \textit{top\_p} of 0.9, and \textit{rep\_penalty} of 1.1. Inference with the GPT-4 model is done using the official OpenAI API, whereas all open-source models are run on 4 Nvidia A100 GPUs.

\subsection{Human study instructions}
\label{app:human_study_details}
Prior to sending each user the instructions we get their consent on how the data will be used through a different form. All annotators are daily english speakers.

\begin{figure*}[htb!]
    \centering
    \includegraphics[width=0.8\linewidth]{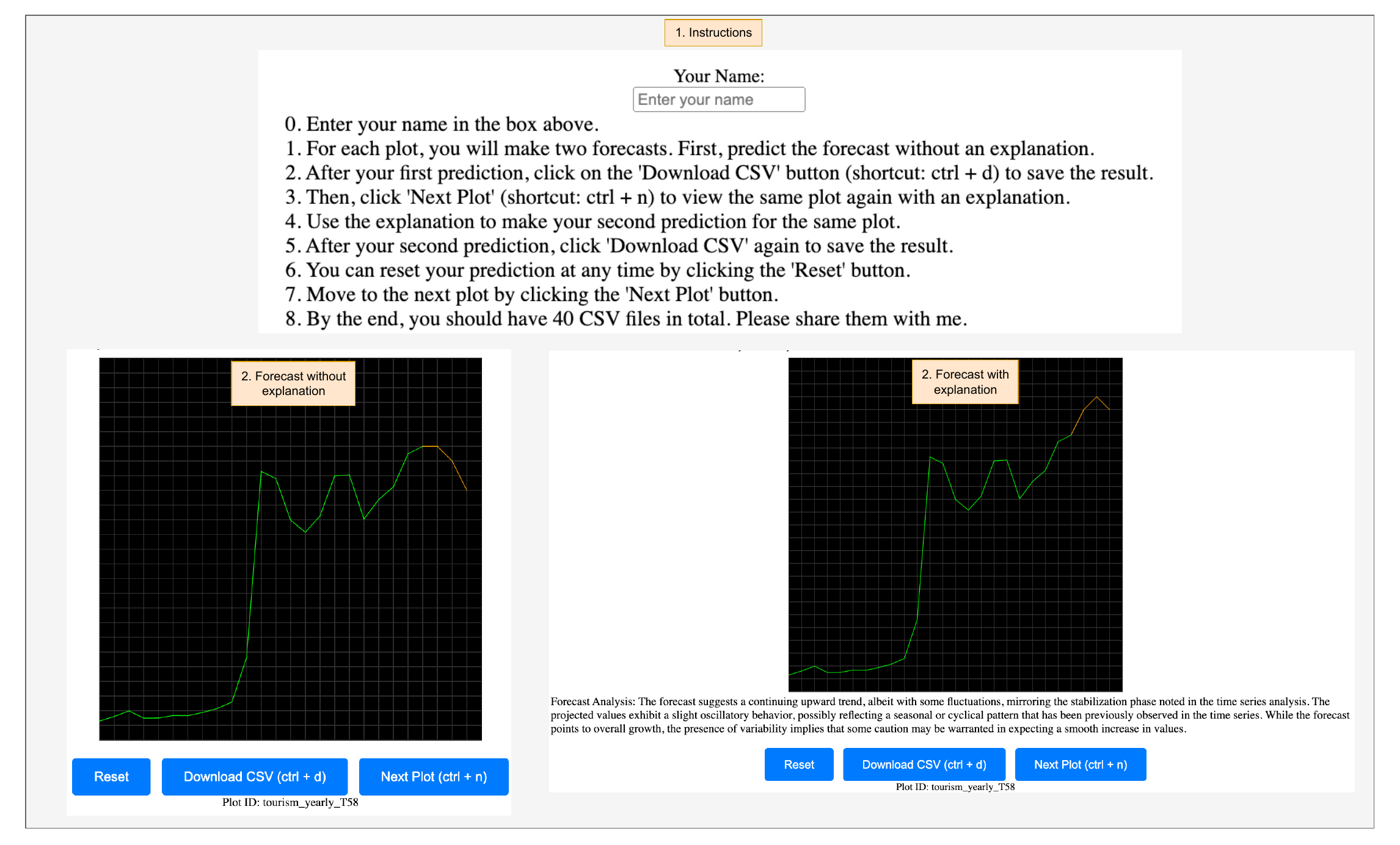}
    \caption{Overview of the human study part 1 setup. (1) Instructions provided to participants. (2) Participants are first asked to draw a forecast without explanation. (3) They then draw the forecast for the same time series using the explanation. This is repated for all 20 time series that we have included in the human study.}
    \label{fig:human_study1}
\end{figure*}

\Cref{fig:human_study1} illustrates the setup for part 1 of the human study. Participants were instructed to generate forecasts for all 20 time series included in the study. For each series, they first drew their forecast without access to any explanation, and then repeated the task after being provided with the explanation.

\begin{figure*}[htb!]
    \centering
    \includegraphics[width=0.8\linewidth]{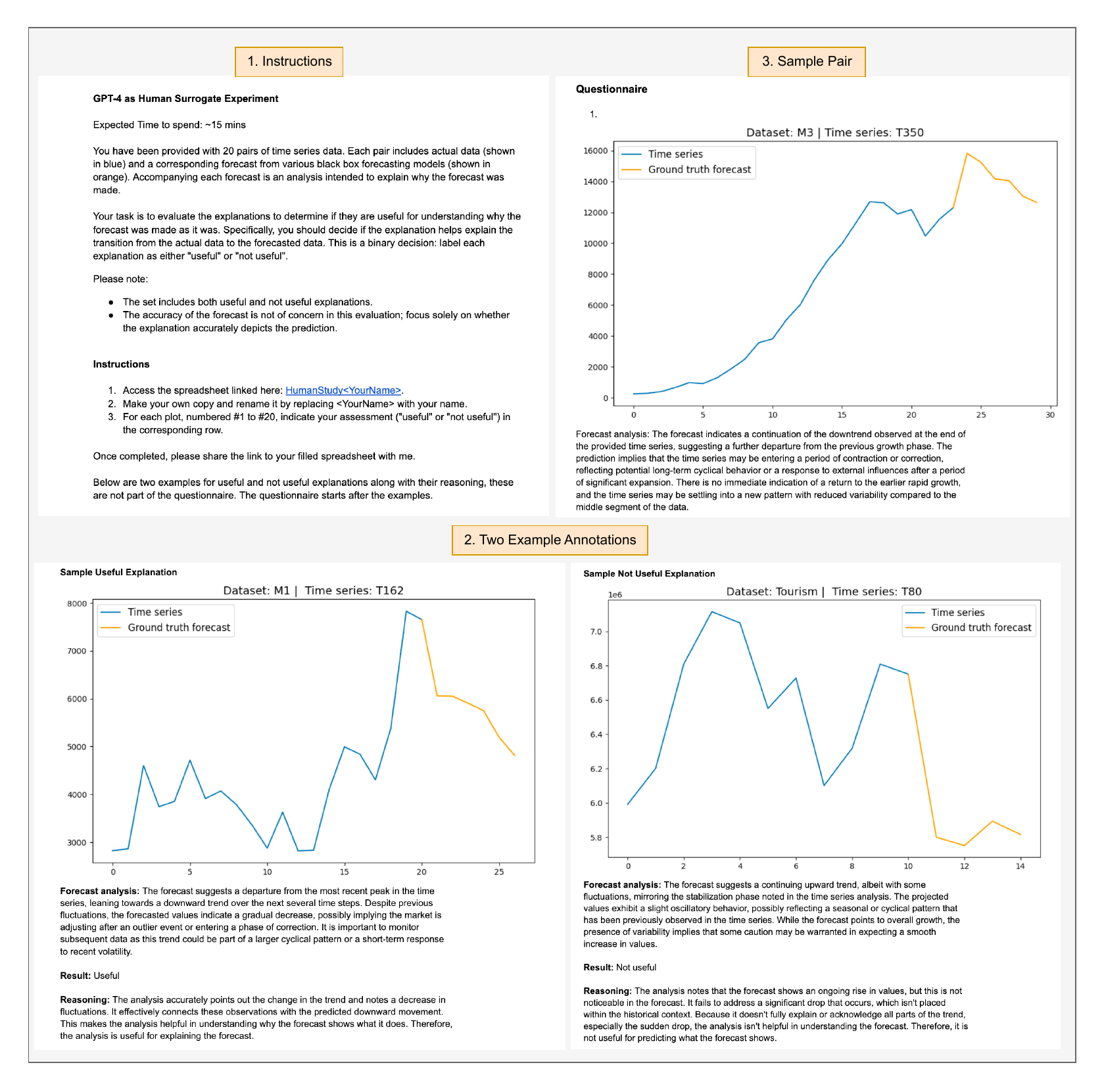}
    \caption{Overview of the human study part 2 setup. (1) Detailed instructions provided to participants. (2) Two example annotations illustrating `useful' and `not useful' explanations along with their reasoning. (3) A sample pair showing a time series forecast explanation and the corresponding forecast, used in the questionnaire.}
    \label{fig:human_study2}
\end{figure*}

\Cref{fig:human_study2} shows the human study part 2 setup with three key components. First, detailed instructions were provided to participants, guiding them on how to evaluate the explanations for their usefulness in understanding the forecasts. Second, two sample annotations illustrated `useful' and `not useful' explanations along with their reasoning. These examples were included to help participants understand the evaluation criteria but were not part of the actual questionnaire. Third, is a sample pair where a time series forecast explanation and the corresponding forecast provided. The users were expected to label 20 such pairs either as `useful' or `not useful'.

% \begin{figure*}[ht!]
%     \centering
%     \begin{subfigure}[t]{\linewidth}
%     \centering
%     \includegraphics[width=0.5\linewidth]{Figures/Grid Simulation_pointer.pdf}
%         \caption{}
%         \label{fig:human_study1_1}
%     \end{subfigure}
%     \quad % Space between the first and second subfigures in the first row
%     \begin{subfigure}[t]{\linewidth}
%     \centering
%     \includegraphics[width=0.5\linewidth]{Figures/Grid Simulation_exp_pointer.pdf}
%         \caption{} 
%         \label{fig:human_study1_2}
%     \end{subfigure}
%     \caption{}
%     \label{fig:human_study1}
% \end{figure*}

\subsection{Additional Results}
\label{app:add_results}
The main paper reported results using the sMAPE distance metric between forecasts. In this section, we share the full results, which also include NMAE and NRMSE metrics, \textit{cf.}~\Cref{tab_app:sim_sanity,tab_app:in_sim_sanity,tab_app:sim_llms,tab_app:in_sim_llms}.

%Sanity check tables
\begin{table*}[!htb]
\resizebox{\textwidth}{!}{%
\begin{tabular}{|c|l|rrr|rrr|rrr|rrr|rrr|}
\hline
\multicolumn{1}{|l|}{}   &                        & \multicolumn{3}{c|}{DeepAR}                                                                                  & \multicolumn{3}{c|}{PatchTST}                                                                                & \multicolumn{3}{c|}{Moirai}                                                                                  & \multicolumn{3}{c|}{ETS}                                                                                     & \multicolumn{3}{c|}{Arima}                                                                                   \\ \hline
Dataset                  & Model                  & \multicolumn{1}{l|}{sMAPE}             & \multicolumn{1}{l|}{NMAE}              & \multicolumn{1}{l|}{NRMSE} & \multicolumn{1}{l|}{sMAPE}             & \multicolumn{1}{l|}{NMAE}              & \multicolumn{1}{l|}{NRMSE} & \multicolumn{1}{l|}{sMAPE}             & \multicolumn{1}{l|}{NMAE}              & \multicolumn{1}{l|}{NRMSE} & \multicolumn{1}{l|}{sMAPE}             & \multicolumn{1}{l|}{NMAE}              & \multicolumn{1}{l|}{NRMSE} & \multicolumn{1}{l|}{\textbf{sMAPE}}    & \multicolumn{1}{l|}{\textbf{NMAE}}     & \multicolumn{1}{l|}{\textbf{NRMSE}} \\ \hline
\multirow{4}{*}{M1}      & LLMTime                & \multicolumn{1}{r|}{4.10E-01}          & \multicolumn{1}{r|}{3.43E-01}          & 3.82E-01                   & \multicolumn{1}{r|}{4.61E-01}          & \multicolumn{1}{r|}{3.99E-01}          & 4.27E-01                   & \multicolumn{1}{r|}{6.87E-01}          & \multicolumn{1}{r|}{9.33E-01}          & 9.81E-01                   & \multicolumn{1}{r|}{3.58E-01}          & \multicolumn{1}{r|}{2.65E-01}          & 2.88E-01                   & \multicolumn{1}{r|}{3.81E-01}          & \multicolumn{1}{r|}{2.92E-01}          & 3.18E-01                            \\ \cline{2-17} 
                         & LLMTime + Random(E)    & \multicolumn{1}{r|}{1.79E-01}          & \multicolumn{1}{r|}{1.82E-01}          & 2.00E-01                   & \multicolumn{1}{r|}{6.62E-01}          & \multicolumn{1}{r|}{6.28E+02}          & 6.29E+02                   & \multicolumn{1}{r|}{4.61E-01}          & \multicolumn{1}{r|}{7.37E-01}          & 7.60E-01                   & \multicolumn{1}{r|}{1.53E-01}          & \multicolumn{1}{r|}{1.62E-01}          & 1.84E-01                   & \multicolumn{1}{r|}{6.01E-01}          & \multicolumn{1}{r|}{6.36E+02}          & 6.38E+02                            \\ \cline{2-17} 
                         & LLMTIME + Monotone (E) & \multicolumn{1}{r|}{2.00E+00}          & \multicolumn{1}{r|}{1.21E+01}          & 1.21E+01                   & \multicolumn{1}{r|}{2.00E+00}          & \multicolumn{1}{r|}{1.25E+01}          & 1.25E+01                   & \multicolumn{1}{r|}{2.00E+00}          & \multicolumn{1}{r|}{1.60E+01}          & 1.60E+01                   & \multicolumn{1}{r|}{2.00E+00}          & \multicolumn{1}{r|}{1.26E+01}          & 1.26E+01                   & \multicolumn{1}{r|}{2.00E+00}          & \multicolumn{1}{r|}{1.21E+01}          & 1.21E+01                            \\ \cline{2-17} 
                         & LLMTime + E            & \multicolumn{1}{r|}{\textbf{1.41E-01}} & \multicolumn{1}{r|}{\textbf{1.41E-01}} & \textbf{1.52E-01}          & \multicolumn{1}{r|}{\textbf{2.10E-01}} & \multicolumn{1}{r|}{\textbf{2.41E-01}} & \textbf{2.67E-01}          & \multicolumn{1}{r|}{\textbf{4.24E-01}} & \multicolumn{1}{r|}{\textbf{6.39E-01}} & \textbf{6.51E-01}          & \multicolumn{1}{r|}{\textbf{9.58E-02}} & \multicolumn{1}{r|}{\textbf{1.08E-01}} & \textbf{1.21E-01}          & \multicolumn{1}{r|}{\textbf{9.21E-02}} & \multicolumn{1}{r|}{\textbf{8.83E-02}} & \textbf{9.69E-02}                   \\ \hline
\multirow{4}{*}{M3}      & LLMTime                & \multicolumn{1}{r|}{4.03E-01}          & \multicolumn{1}{r|}{3.06E-01}          & 3.26E-01                   & \multicolumn{1}{r|}{4.44E-01}          & \multicolumn{1}{r|}{3.61E-01}          & 4.07E-01                   & \multicolumn{1}{r|}{5.47E-01}          & \multicolumn{1}{r|}{5.67E-01}          & 5.94E-01                   & \multicolumn{1}{r|}{3.45E-01}          & \multicolumn{1}{r|}{2.55E-01}          & 2.81E-01                   & \multicolumn{1}{r|}{3.50E-01}          & \multicolumn{1}{r|}{2.55E-01}          & 2.78E-01                            \\ \cline{2-17} 
                         & LLMTime + Random(E)    & \multicolumn{1}{r|}{2.55E-01}          & \multicolumn{1}{r|}{2.29E-01}          & 2.47E-01                   & \multicolumn{1}{r|}{2.41E-01}          & \multicolumn{1}{r|}{2.27E-01}          & 2.57E-01                   & \multicolumn{1}{r|}{3.49E-01}          & \multicolumn{1}{r|}{4.23E-01}          & 4.43E-01                   & \multicolumn{1}{r|}{1.49E-01}          & \multicolumn{1}{r|}{1.36E-01}          & 1.53E-01                   & \multicolumn{1}{r|}{1.52E-01}          & \multicolumn{1}{r|}{1.45E-01}          & 1.64E-01                            \\ \cline{2-17} 
                         & LLMTIME + Monotone (E) & \multicolumn{1}{r|}{2.00E+00}          & \multicolumn{1}{r|}{1.42E+01}          & 1.42E+01                   & \multicolumn{1}{r|}{2.00E+00}          & \multicolumn{1}{r|}{1.39E+01}          & 1.39E+01                   & \multicolumn{1}{r|}{2.00E+00}          & \multicolumn{1}{r|}{1.55E+01}          & 1.55E+01                   & \multicolumn{1}{r|}{2.00E+00}          & \multicolumn{1}{r|}{1.43E+01}          & 1.43E+01                   & \multicolumn{1}{r|}{2.00E+00}          & \multicolumn{1}{r|}{1.49E+01}          & 1.49E+01                            \\ \cline{2-17} 
                         & LLMTime + E            & \multicolumn{1}{r|}{\textbf{1.86E-01}} & \multicolumn{1}{r|}{\textbf{1.83E-01}} & \textbf{1.96E-01}          & \multicolumn{1}{r|}{\textbf{2.08E-01}} & \multicolumn{1}{r|}{\textbf{2.11E-01}} & \textbf{2.38E-01}          & \multicolumn{1}{r|}{\textbf{3.17E-01}} & \multicolumn{1}{r|}{\textbf{4.08E-01}} & \textbf{4.20E-01}          & \multicolumn{1}{r|}{\textbf{7.57E-02}} & \multicolumn{1}{r|}{\textbf{9.78E-02}} & \textbf{1.05E-01}          & \multicolumn{1}{r|}{\textbf{8.81E-02}} & \multicolumn{1}{r|}{\textbf{8.77E-02}} & \textbf{9.49E-02}                   \\ \hline
\multirow{4}{*}{Tourism} & LLMTime                & \multicolumn{1}{r|}{4.87E-01}          & \multicolumn{1}{r|}{4.43E-01}          & 4.68E-01                   & \multicolumn{1}{r|}{5.45E-01}          & \multicolumn{1}{r|}{5.68E-01}          & 6.06E-01                   & \multicolumn{1}{r|}{6.53E-01}          & \multicolumn{1}{r|}{7.06E-01}          & 7.28E-01                   & \multicolumn{1}{r|}{4.16E-01}          & \multicolumn{1}{r|}{3.42E-01}          & 3.68E-01                   & \multicolumn{1}{r|}{4.07E-01}          & \multicolumn{1}{r|}{3.29E-01}          & 3.56E-01                            \\ \cline{2-17} 
                         & LLMTime + Random(E)    & \multicolumn{1}{r|}{2.87E-01}          & \multicolumn{1}{r|}{4.71E-01}          & 5.07E-01                   & \multicolumn{1}{r|}{3.36E-01}          & \multicolumn{1}{r|}{4.23E-01}          & 4.60E-01                   & \multicolumn{1}{r|}{3.88E-01}          & \multicolumn{1}{r|}{\textbf{5.46E-01}} & \textbf{5.65E-01}          & \multicolumn{1}{r|}{1.62E-01}          & \multicolumn{1}{r|}{1.70E-01}          & 1.88E-01                   & \multicolumn{1}{r|}{1.67E-01}          & \multicolumn{1}{r|}{1.89E-01}          & 2.09E-01                            \\ \cline{2-17} 
                         & LLMTIME + Monotone (E) & \multicolumn{1}{r|}{2.00E+00}          & \multicolumn{1}{r|}{1.78E+01}          & 1.78E+01                   & \multicolumn{1}{r|}{2.00E+00}          & \multicolumn{1}{r|}{1.56E+01}          & 1.56E+01                   & \multicolumn{1}{r|}{2.00E+00}          & \multicolumn{1}{r|}{1.62E+01}          & 1.62E+01                   & \multicolumn{1}{r|}{2.00E+00}          & \multicolumn{1}{r|}{1.31E+01}          & 1.31E+01                   & \multicolumn{1}{r|}{2.00E+00}          & \multicolumn{1}{r|}{1.38E+01}          & 1.38E+01                            \\ \cline{2-17} 
                         & LLMTime + E            & \multicolumn{1}{r|}{\textbf{2.05E-01}} & \multicolumn{1}{r|}{\textbf{2.42E-01}} & \textbf{2.54E-01}          & \multicolumn{1}{r|}{\textbf{2.96E-01}} & \multicolumn{1}{r|}{\textbf{3.76E-01}} & \textbf{4.06E-01}          & \multicolumn{1}{r|}{\textbf{3.87E-01}} & \multicolumn{1}{r|}{5.66E-01}          & 5.75E-01                   & \multicolumn{1}{r|}{\textbf{8.80E-02}} & \multicolumn{1}{r|}{\textbf{9.58E-02}} & \textbf{1.01E-01}          & \multicolumn{1}{r|}{\textbf{7.10E-02}} & \multicolumn{1}{r|}{\textbf{7.62E-02}} & \textbf{8.17E-02}                   \\ \hline
\end{tabular}%
}
\caption{Sanity check experiments for simulatability. We report NMAE and NRMSE additional to sMAPE from the main paper results. Results are averaged over three runs, lower is better.}
\label{tab_app:sim_sanity}
\end{table*}
\begin{table*}[!htb]
\resizebox{\textwidth}{!}{%
\begin{tabular}{|c|l|rrr|rrr|rrr|rrr|rrr|rrr|}
\hline
\multicolumn{1}{|l|}{}            &                        & \multicolumn{3}{c|}{\textbf{DeepAR}}                                                                                  & \multicolumn{3}{c|}{\textbf{PatchTST}}                                                                                & \multicolumn{3}{c|}{\textbf{Auto Arima}}                                                                              & \multicolumn{3}{c|}{\textbf{Moirai}}                                                                                  & \multicolumn{3}{c|}{\textbf{ETS}}                                                                                     & \multicolumn{3}{c|}{\textbf{Arima}}                                                                                   \\ \hline
\textbf{Dataset}                  & \textbf{Model}         & \multicolumn{1}{l|}{\textbf{sMAPE}}    & \multicolumn{1}{l|}{\textbf{NMAE}}     & \multicolumn{1}{l|}{\textbf{NRMSE}} & \multicolumn{1}{l|}{\textbf{sMAPE}}    & \multicolumn{1}{l|}{\textbf{NMAE}}     & \multicolumn{1}{l|}{\textbf{NRMSE}} & \multicolumn{1}{l|}{\textbf{sMAPE}}    & \multicolumn{1}{l|}{\textbf{NMAE}}     & \multicolumn{1}{l|}{\textbf{NRMSE}} & \multicolumn{1}{l|}{\textbf{sMAPE}}    & \multicolumn{1}{l|}{\textbf{NMAE}}     & \multicolumn{1}{l|}{\textbf{NRMSE}} & \multicolumn{1}{l|}{\textbf{sMAPE}}    & \multicolumn{1}{l|}{\textbf{NMAE}}     & \multicolumn{1}{l|}{\textbf{NRMSE}} & \multicolumn{1}{l|}{\textbf{sMAPE}}    & \multicolumn{1}{l|}{\textbf{NMAE}}     & \multicolumn{1}{l|}{\textbf{NRMSE}} \\ \hline
\multirow{4}{*}{\textbf{M1}}      & LLMTime                & \multicolumn{1}{r|}{3.85E-01}          & \multicolumn{1}{r|}{2.20E+02}          & 4.43E+02                            & \multicolumn{1}{r|}{4.25E-01}          & \multicolumn{1}{r|}{1.16E+01}          & 2.37E+01                            & \multicolumn{1}{r|}{3.46E-01}          & \multicolumn{1}{r|}{5.09E-01}          & 5.63E-01                            & \multicolumn{1}{r|}{6.65E-01}          & \multicolumn{1}{r|}{1.09E+00}          & 1.15E+00                            & \multicolumn{1}{r|}{3.24E-01}          & \multicolumn{1}{r|}{4.90E-01}          & 5.50E-01                            & \multicolumn{1}{r|}{3.46E-01}          & \multicolumn{1}{r|}{5.09E-01}          & 5.63E-01                            \\ \cline{2-20} 
                                  & LLMTime + Random(E)    & \multicolumn{1}{r|}{3.27E-01}          & \multicolumn{1}{r|}{8.71E+01}          & 1.84E+02                            & \multicolumn{1}{r|}{3.62E-01}          & \multicolumn{1}{r|}{9.33E+00}          & 1.85E+01                            & \multicolumn{1}{r|}{2.74E-01}          & \multicolumn{1}{r|}{5.55E-01}          & 6.24E-01                            & \multicolumn{1}{r|}{5.76E-01}          & \multicolumn{1}{r|}{1.05E+00}          & 1.10E+00                            & \multicolumn{1}{r|}{2.43E-01}          & \multicolumn{1}{r|}{4.99E-01}          & 5.66E-01                            & \multicolumn{1}{r|}{2.74E-01}          & \multicolumn{1}{r|}{5.55E-01}          & 6.24E-01                            \\ \cline{2-20} 
                                  & LLMTIME + Monotone (E) & \multicolumn{1}{r|}{1.98E+00}          & \multicolumn{1}{r|}{1.68E+01}          & 1.68E+01                            & \multicolumn{1}{r|}{1.98E+00}          & \multicolumn{1}{r|}{1.44E+01}          & 1.44E+01                            & \multicolumn{1}{r|}{1.99E+00}          & \multicolumn{1}{r|}{2.41E+01}          & 2.41E+01                            & \multicolumn{1}{r|}{1.98E+00}          & \multicolumn{1}{r|}{2.03E+01}          & 2.03E+01                            & \multicolumn{1}{r|}{2.00E+00}          & \multicolumn{1}{r|}{1.91E+01}          & 1.91E+01                            & \multicolumn{1}{r|}{1.99E+00}          & \multicolumn{1}{r|}{2.41E+01}          & 2.41E+01                            \\ \cline{2-20} 
                                  & LLMTime + E            & \multicolumn{1}{r|}{\textbf{2.97E-01}} & \multicolumn{1}{r|}{\textbf{6.81E+00}} & \textbf{8.49E+00}                   & \multicolumn{1}{r|}{\textbf{3.26E-01}} & \multicolumn{1}{r|}{\textbf{8.70E-01}} & \textbf{1.02E+00}                   & \multicolumn{1}{r|}{\textbf{2.52E-01}} & \multicolumn{1}{r|}{\textbf{4.52E-01}} & \textbf{4.88E-01}                   & \multicolumn{1}{r|}{\textbf{5.58E-01}} & \multicolumn{1}{r|}{\textbf{1.01E+00}} & \textbf{1.05E+00}                   & \multicolumn{1}{r|}{\textbf{2.02E-01}} & \multicolumn{1}{r|}{\textbf{4.56E-01}} & \textbf{5.08E-01}                   & \multicolumn{1}{r|}{\textbf{2.52E-01}} & \multicolumn{1}{r|}{\textbf{4.52E-01}} & \textbf{4.88E-01}                   \\ \hline
\multirow{4}{*}{\textbf{M3}}      & LLMTime                & \multicolumn{1}{r|}{4.16E-01}          & \multicolumn{1}{r|}{5.27E-01}          & 5.82E-01                            & \multicolumn{1}{r|}{5.10E-01}          & \multicolumn{1}{r|}{5.38E-01}          & 6.09E-01                            & \multicolumn{1}{r|}{4.67E-01}          & \multicolumn{1}{r|}{1.07E+00}          & 1.23E+00                            & \multicolumn{1}{r|}{7.30E-01}          & \multicolumn{1}{r|}{1.47E+00}          & 1.61E+00                            & \multicolumn{1}{r|}{5.26E-01}          & \multicolumn{1}{r|}{1.69E+00}          & 1.88E+00                            & \multicolumn{1}{r|}{4.67E-01}          & \multicolumn{1}{r|}{1.07E+00}          & 1.23E+00                            \\ \cline{2-20} 
                                  & LLMTime + Random(E)    & \multicolumn{1}{r|}{3.69E-01}          & \multicolumn{1}{r|}{5.11E-01}          & 5.70E-01                            & \multicolumn{1}{r|}{4.29E-01}          & \multicolumn{1}{r|}{5.19E-01}          & 5.88E-01                            & \multicolumn{1}{r|}{4.35E-01}          & \multicolumn{1}{r|}{9.14E-01}          & 1.05E+00                            & \multicolumn{1}{r|}{5.99E-01}          & \multicolumn{1}{r|}{1.29E+00}          & 1.41E+00                            & \multicolumn{1}{r|}{4.47E-01}          & \multicolumn{1}{r|}{1.51E+00}          & 1.68E+00                            & \multicolumn{1}{r|}{4.35E-01}          & \multicolumn{1}{r|}{9.14E-01}          & 1.05E+00                            \\ \cline{2-20} 
                                  & LLMTIME + Monotone (E) & \multicolumn{1}{r|}{1.97E+00}          & \multicolumn{1}{r|}{4.06E+01}          & 4.06E+01                            & \multicolumn{1}{r|}{1.98E+00}          & \multicolumn{1}{r|}{1.83E+01}          & 1.83E+01                            & \multicolumn{1}{r|}{1.98E+00}          & \multicolumn{1}{r|}{3.82E+01}          & 3.82E+01                            & \multicolumn{1}{r|}{1.98E+00}          & \multicolumn{1}{r|}{2.30E+01}          & 2.30E+01                            & \multicolumn{1}{r|}{1.99E+00}          & \multicolumn{1}{r|}{4.20E+01}          & 4.55E+01                            & \multicolumn{1}{r|}{1.98E+00}          & \multicolumn{1}{r|}{3.82E+01}          & 3.82E+01                            \\ \cline{2-20} 
                                  & LLMTime + E            & \multicolumn{1}{r|}{\textbf{3.28E-01}} & \multicolumn{1}{r|}{\textbf{4.84E-01}} & \textbf{5.31E-01}                   & \multicolumn{1}{r|}{\textbf{3.87E-01}} & \multicolumn{1}{r|}{\textbf{4.34E-01}} & \textbf{4.99E-01}                   & \multicolumn{1}{r|}{\textbf{3.93E-01}} & \multicolumn{1}{r|}{\textbf{7.32E-01}} & \textbf{8.13E-01}                   & \multicolumn{1}{r|}{\textbf{5.87E-01}} & \multicolumn{1}{r|}{\textbf{1.25E+00}} & \textbf{1.37E+00}                   & \multicolumn{1}{r|}{\textbf{3.99E-01}} & \multicolumn{1}{r|}{\textbf{1.34E+00}} & \textbf{1.41E+00}                   & \multicolumn{1}{r|}{\textbf{3.93E-01}} & \multicolumn{1}{r|}{\textbf{7.32E-01}} & \textbf{8.13E-01}                   \\ \hline
\multirow{4}{*}{\textbf{Tourism}} & LLMTime                & \multicolumn{1}{r|}{4.55E-01}          & \multicolumn{1}{r|}{6.08E-01}          & 6.64E-01                            & \multicolumn{1}{r|}{4.76E-01}          & \multicolumn{1}{r|}{2.04E+00}          & 2.56E+00                            & \multicolumn{1}{r|}{4.51E-01}          & \multicolumn{1}{r|}{3.95E-01}          & 4.43E-01                            & \multicolumn{1}{r|}{6.66E-01}          & \multicolumn{1}{r|}{2.93E+00}          & 3.05E+00                            & \multicolumn{1}{r|}{4.16E-01}          & \multicolumn{1}{r|}{4.47E+00}          & 4.67E+00                            & \multicolumn{1}{r|}{4.51E-01}          & \multicolumn{1}{r|}{3.95E-01}          & 4.43E-01                            \\ \cline{2-20} 
                                  & LLMTime + Random(E)    & \multicolumn{1}{r|}{3.41E-01}          & \multicolumn{1}{r|}{4.87E-01}          & 5.44E-01                            & \multicolumn{1}{r|}{3.77E-01}          & \multicolumn{1}{r|}{3.18E+00}          & 4.96E+00                            & \multicolumn{1}{r|}{4.13E-01}          & \multicolumn{1}{r|}{4.01E-01}          & 4.56E-01                            & \multicolumn{1}{r|}{5.36E-01}          & \multicolumn{1}{r|}{2.95E+00}          & 3.07E+00                            & \multicolumn{1}{r|}{3.62E-01}          & \multicolumn{1}{r|}{4.58E+00}          & 4.81E+00                            & \multicolumn{1}{r|}{4.13E-01}          & \multicolumn{1}{r|}{4.01E-01}          & 4.56E-01                            \\ \cline{2-20} 
                                  & LLMTIME + Monotone (E) & \multicolumn{1}{r|}{1.44E+00}          & \multicolumn{1}{r|}{1.80E+01}          & 1.81E+01                            & \multicolumn{1}{r|}{1.98E+00}          & \multicolumn{1}{r|}{1.61E+01}          & 1.61E+01                            & \multicolumn{1}{r|}{1.99E+00}          & \multicolumn{1}{r|}{2.23E+01}          & 2.23E+01                            & \multicolumn{1}{r|}{1.99E+00}          & \multicolumn{1}{r|}{3.51E+01}          & 3.51E+01                            & \multicolumn{1}{r|}{1.99E+00}          & \multicolumn{1}{r|}{7.09E+01}          & 7.09E+01                            & \multicolumn{1}{r|}{1.99E+00}          & \multicolumn{1}{r|}{2.23E+01}          & 2.23E+01                            \\ \cline{2-20} 
                                  & LLMTime + E            & \multicolumn{1}{r|}{\textbf{2.91E-01}} & \multicolumn{1}{r|}{\textbf{4.26E-01}} & \textbf{4.76E-01}                   & \multicolumn{1}{r|}{\textbf{3.35E-01}} & \multicolumn{1}{r|}{\textbf{1.79E+00}} & \textbf{2.28E+00}                   & \multicolumn{1}{r|}{\textbf{3.76E-01}} & \multicolumn{1}{r|}{\textbf{3.27E-01}} & \textbf{3.56E-01}                   & \multicolumn{1}{r|}{\textbf{5.23E-01}} & \multicolumn{1}{r|}{\textbf{2.85E+00}} & \textbf{2.94E+00}                   & \multicolumn{1}{r|}{\textbf{2.96E-01}} & \multicolumn{1}{r|}{\textbf{3.84E+00}} & \textbf{3.96E+00}                   & \multicolumn{1}{r|}{\textbf{3.76E-01}} & \multicolumn{1}{r|}{\textbf{3.27E-01}} & \textbf{3.56E-01}                   \\ \hline
\end{tabular}
}
\caption{Sanity check experiments for synthetic simulatability. We report NMAE and NRMSE additional to sMAPE from the main paper results. Results are averaged over three runs, lower is better.}
\label{tab_app:in_sim_sanity}
\end{table*}

%OSS models
\begin{table*}[!htb]
\resizebox{\textwidth}{!}{%
\begin{tabular}{|c|l|rrr|rrr|rrr|rrr|rrr|}
\hline
\multicolumn{1}{|l|}{}                    &                          & \multicolumn{3}{c|}{\textbf{DeepAR}}                                                                                  & \multicolumn{3}{c|}{\textbf{PatchTST}}                                                                                & \multicolumn{3}{c|}{\textbf{Moirai}}                                                                                  & \multicolumn{3}{c|}{\textbf{ETS}}                                                                                & \multicolumn{3}{c|}{\textbf{Arima}}                                                                              \\ \hline
\multicolumn{1}{|l|}{\textbf{Dataset}}    & \textbf{Model}           & \multicolumn{1}{l|}{\textbf{sMAPE}}    & \multicolumn{1}{l|}{\textbf{NMAE}}     & \multicolumn{1}{l|}{\textbf{NRMSE}} & \multicolumn{1}{l|}{\textbf{sMAPE}}    & \multicolumn{1}{l|}{\textbf{NMAE}}     & \multicolumn{1}{l|}{\textbf{NRMSE}} & \multicolumn{1}{l|}{\textbf{sMAPE}}    & \multicolumn{1}{l|}{\textbf{NMAE}}     & \multicolumn{1}{l|}{\textbf{NRMSE}} & \multicolumn{1}{l|}{\textbf{sMAPE}}    & \multicolumn{1}{l|}{\textbf{NMAE}}     & \multicolumn{1}{l|}{\textbf{NRMSE}} & \multicolumn{1}{l|}{\textbf{sMAPE}}    & \multicolumn{1}{l|}{\textbf{NMAE}}     & \multicolumn{1}{l|}{\textbf{NRMSE}} \\ \hline
\multirow{5}{*}{\textbf{M1}}              & LLMTime + E (GPT4)       & \multicolumn{1}{r|}{\textbf{1.47E-01}} & \multicolumn{1}{r|}{\textbf{1.40E-01}} & \textbf{1.62E-01}                   & \multicolumn{1}{r|}{\textbf{2.02E-01}} & \multicolumn{1}{r|}{\textbf{2.21E-01}} & \textbf{2.47E-01}                   & \multicolumn{1}{r|}{\textbf{4.06E-01}} & \multicolumn{1}{r|}{\textbf{6.01E-01}} & \textbf{6.14E-01}                   & \multicolumn{1}{r|}{\textbf{9.80E-02}} & \multicolumn{1}{r|}{\textbf{9.70E-02}} & \textbf{1.08E-01}                   & \multicolumn{1}{r|}{1.03E-01}          & \multicolumn{1}{r|}{9.63E-02}          & \textbf{1.06E-01}                   \\ \cline{2-17} 
                                          & LLMTime + E (Llama3-70b) & \multicolumn{1}{r|}{1.65E-01}          & \multicolumn{1}{r|}{1.71E-01}          & 1.88E-01                            & \multicolumn{1}{r|}{2.61E-01}          & \multicolumn{1}{r|}{4.64E-01}          & 5.45E-01                            & \multicolumn{1}{r|}{4.25E-01}          & \multicolumn{1}{r|}{6.60E-01}          & 6.75E-01                            & \multicolumn{1}{r|}{1.00E-01}          & \multicolumn{1}{r|}{1.22E-01}          & 1.42E-01                            & \multicolumn{1}{r|}{\textbf{8.90E-02}} & \multicolumn{1}{r|}{9.46E-02}          & 1.08E-01                            \\ \cline{2-17} 
                                          & LLMTime + E (Llama2-70b) & \multicolumn{1}{r|}{4.72E-01}          & \multicolumn{1}{r|}{8.98E-01}          & 1.58E+00                            & \multicolumn{1}{r|}{4.35E-01}          & \multicolumn{1}{r|}{3.49E-01}          & 3.82E-01                            & \multicolumn{1}{r|}{5.93E-01}          & \multicolumn{1}{r|}{5.94E-01}          & 6.21E-01                            & \multicolumn{1}{r|}{3.99E-01}          & \multicolumn{1}{r|}{2.95E-01}          & 3.38E-01                            & \multicolumn{1}{r|}{3.80E-01}          & \multicolumn{1}{r|}{2.99E-01}          & 3.28E-01                            \\ \cline{2-17} 
                                          & LLMTime + E (Vicuna-7b)  & \multicolumn{1}{r|}{2.92E-01}          & \multicolumn{1}{r|}{2.61E-01}          & 2.86E-01                            & \multicolumn{1}{r|}{2.94E-01}          & \multicolumn{1}{r|}{2.94E-01}          & 3.22E-01                            & \multicolumn{1}{r|}{5.92E-01}          & \multicolumn{1}{r|}{8.05E-01}          & 8.23E-01                            & \multicolumn{1}{r|}{2.07E-01}          & \multicolumn{1}{r|}{1.88E-01}          & 2.08E-01                            & \multicolumn{1}{r|}{2.48E-01}          & \multicolumn{1}{r|}{2.33E-01}          & 2.52E-01                            \\ \cline{2-17} 
                                          & LLMTime + E (Mistral-7b) & \multicolumn{1}{r|}{1.62E-01}          & \multicolumn{1}{r|}{1.63E-01}          & 1.77E-01                            & \multicolumn{1}{r|}{2.46E-01}          & \multicolumn{1}{r|}{2.98E-01}          & 3.26E-01                            & \multicolumn{1}{r|}{5.04E-01}          & \multicolumn{1}{r|}{1.18E+00}          & 1.20E+00                            & \multicolumn{1}{r|}{2.00E-01}          & \multicolumn{1}{r|}{2.91E+00}          & 2.93E+00                            & \multicolumn{1}{r|}{1.66E-01}          & \multicolumn{1}{r|}{1.90E-01}          & 2.07E-01                            \\ \hline
\multicolumn{1}{|l|}{\multirow{5}{*}{M3}} & LLMTime + E (GPT4)       & \multicolumn{1}{r|}{\textbf{1.83E-01}} & \multicolumn{1}{r|}{\textbf{1.78E-01}} & \textbf{1.89E-01}                   & \multicolumn{1}{r|}{\textbf{2.09E-01}} & \multicolumn{1}{r|}{\textbf{2.14E-01}} & \textbf{2.42E-01}                   & \multicolumn{1}{r|}{\textbf{2.88E-01}} & \multicolumn{1}{r|}{\textbf{3.46E-01}} & \textbf{3.59E-01}                   & \multicolumn{1}{r|}{\textbf{6.83E-02}} & \multicolumn{1}{r|}{\textbf{9.25E-02}} & \textbf{9.89E-02}                   & \multicolumn{1}{r|}{\textbf{8.54E-02}} & \multicolumn{1}{r|}{1.45E-01}          & 1.54E-01                            \\ \cline{2-17} 
\multicolumn{1}{|l|}{}                    & LLMTime + E (Llama3-70b) & \multicolumn{1}{r|}{2.13E-01}          & \multicolumn{1}{r|}{2.21E-01}          & 2.39E-01                            & \multicolumn{1}{r|}{2.34E-01}          & \multicolumn{1}{r|}{2.37E-01}          & 2.65E-01                            & \multicolumn{1}{r|}{2.98E-01}          & \multicolumn{1}{r|}{4.03E-01}          & 4.23E-01                            & \multicolumn{1}{r|}{1.14E-01}          & \multicolumn{1}{r|}{1.09E-01}          & 1.21E-01                            & \multicolumn{1}{r|}{1.03E-01}          & \multicolumn{1}{r|}{\textbf{1.10E-01}} & \textbf{1.22E-01}                   \\ \cline{2-17} 
\multicolumn{1}{|l|}{}                    & LLMTime + E (Llama2-70b) & \multicolumn{1}{r|}{3.65E-01}          & \multicolumn{1}{r|}{2.78E-01}          & 2.95E-01                            & \multicolumn{1}{r|}{3.61E-01}          & \multicolumn{1}{r|}{2.89E-01}          & 3.23E-01                            & \multicolumn{1}{r|}{5.11E-01}          & \multicolumn{1}{r|}{5.80E-01}          & 6.03E-01                            & \multicolumn{1}{r|}{3.07E-01}          & \multicolumn{1}{r|}{2.87E-01}          & 3.26E-01                            & \multicolumn{1}{r|}{3.20E-01}          & \multicolumn{1}{r|}{2.88E-01}          & 3.17E-01                            \\ \cline{2-17} 
\multicolumn{1}{|l|}{}                    & LLMTime + E (Vicuna-7b)  & \multicolumn{1}{r|}{2.96E-01}          & \multicolumn{1}{r|}{2.79E-01}          & 2.98E-01                            & \multicolumn{1}{r|}{2.80E-01}          & \multicolumn{1}{r|}{2.71E-01}          & 3.02E-01                            & \multicolumn{1}{r|}{4.19E-01}          & \multicolumn{1}{r|}{5.04E-01}          & 5.21E-01                            & \multicolumn{1}{r|}{1.76E-01}          & \multicolumn{1}{r|}{1.92E-01}          & 2.08E-01                            & \multicolumn{1}{r|}{2.11E-01}          & \multicolumn{1}{r|}{2.31E-01}          & 2.49E-01                            \\ \cline{2-17} 
\multicolumn{1}{|l|}{}                    & LLMTime + E (Mistral-7b) & \multicolumn{1}{r|}{2.21E-01} & \multicolumn{1}{r|}{\textbf{3.18E-01}} & \textbf{3.45E-01}                   & \multicolumn{1}{r|}{2.70E-01}          & \multicolumn{1}{r|}{4.89E-01}          & 5.38E-01                            & \multicolumn{1}{r|}{3.74E-01}          & \multicolumn{1}{r|}{5.87E-01}          & 6.07E-01                            & \multicolumn{1}{r|}{2.45E-01}          & \multicolumn{1}{r|}{7.26E-01}          & 7.49E-01                            & \multicolumn{1}{r|}{2.22E-01} & \multicolumn{1}{r|}{6.37E-01}          & 6.61E-01                            \\ \hline
\multirow{5}{*}{\textbf{Tourism}}         & LLMTime + E (GPT4)       & \multicolumn{1}{r|}{\textbf{1.98E-01}} & \multicolumn{1}{r|}{\textbf{2.37E-01}} & \textbf{2.50E-01}                   & \multicolumn{1}{r|}{\textbf{2.93E-01}} & \multicolumn{1}{r|}{\textbf{3.79E-01}} & \textbf{4.10E-01}                   & \multicolumn{1}{r|}{\textbf{3.92E-01}} & \multicolumn{1}{r|}{\textbf{5.77E-01}} & \textbf{5.88E-01}                   & \multicolumn{1}{r|}{\textbf{9.04E-02}} & \multicolumn{1}{r|}{\textbf{9.59E-02}} & \textbf{1.02E-01}                   & \multicolumn{1}{r|}{\textbf{6.71E-02}} & \multicolumn{1}{r|}{\textbf{6.73E-02}} & \textbf{7.26E-02}                   \\ \cline{2-17} 
                                          & LLMTime + E (Llama3-70b) & \multicolumn{1}{r|}{2.41E-01}          & \multicolumn{1}{r|}{3.64E-01}          & 3.86E-01                            & \multicolumn{1}{r|}{3.31E-01}          & \multicolumn{1}{r|}{4.32E-01}          & 4.65E-01                            & \multicolumn{1}{r|}{4.14E-01}          & \multicolumn{1}{r|}{6.40E-01}          & 6.50E-01                            & \multicolumn{1}{r|}{1.49E-01}          & \multicolumn{1}{r|}{1.47E-01}          & 1.56E-01                            & \multicolumn{1}{r|}{1.15E-01}          & \multicolumn{1}{r|}{1.35E-01}          & 1.47E-01                            \\ \cline{2-17} 
                                          & LLMTime + E (Llama2-70b) & \multicolumn{1}{r|}{3.73E-01} & \multicolumn{1}{r|}{4.25E-01} & 4.49E-01                 & \multicolumn{1}{r|}{4.63E-01}          & \multicolumn{1}{r|}{4.96E-01}          & 5.32E-01                            & \multicolumn{1}{r|}{5.53E-01}          & \multicolumn{1}{r|}{6.65E-01}          & 6.82E-01                            & \multicolumn{1}{r|}{3.39E-01}          & \multicolumn{1}{r|}{2.76E-01}          & 2.92E-01                            & \multicolumn{1}{r|}{3.34E-01}          & \multicolumn{1}{r|}{3.13E-01} & 3.36E-01                   \\ \cline{2-17} 
                                          & LLMTime + E (Vicuna-7b)  & \multicolumn{1}{r|}{3.15E-01}          & \multicolumn{1}{r|}{4.72E-01}          & 4.88E-01                            & \multicolumn{1}{r|}{3.81E-01}          & \multicolumn{1}{r|}{4.56E-01}          & 4.88E-01                            & \multicolumn{1}{r|}{4.96E-01}          & \multicolumn{1}{r|}{6.89E-01}          & 7.04E-01                            & \multicolumn{1}{r|}{2.15E-01}          & \multicolumn{1}{r|}{9.53E-01}          & 1.09E+00                            & \multicolumn{1}{r|}{1.91E-01}          & \multicolumn{1}{r|}{4.05E-01}          & 4.30E-01                            \\ \cline{2-17} 
                                          & LLMTime + E (Mistral-7b) & \multicolumn{1}{r|}{2.83E-01}          & \multicolumn{1}{r|}{2.67E+01}          & 2.72E+01                            & \multicolumn{1}{r|}{3.88E-01}          & \multicolumn{1}{r|}{3.65E+00}          & 3.74E+00                            & \multicolumn{1}{r|}{4.71E-01}          & \multicolumn{1}{r|}{1.23E+01}          & 1.24E+01                            & \multicolumn{1}{r|}{2.53E-01}          & \multicolumn{1}{r|}{3.48E+00}          & 3.82E+00                            & \multicolumn{1}{r|}{2.53E-01}          & \multicolumn{1}{r|}{2.45E+00}          & 2.48E+00                            \\ \hline
\end{tabular}
}
\caption{Performance comparison across different LLMs as baseline explainer backbone for simulatability. We report NMAE and NRMSE additional to sMAPE
from the main paper results, lower is better.}
\label{tab_app:sim_llms}
\end{table*}
\begin{table*}[!htb]
\resizebox{\textwidth}{!}{%
\begin{tabular}{|c|l|rrr|rrr|rrr|rrr|rrr|}
\hline
\multicolumn{1}{|l|}{}                    &                          & \multicolumn{3}{c|}{\textbf{DeepAR}}                                                                           & \multicolumn{3}{c|}{\textbf{PatchTST}}                                                                         & \multicolumn{3}{c|}{\textbf{Moirai}}                                                                           & \multicolumn{3}{c|}{\textbf{ETS}}                                                                         & \multicolumn{3}{c|}{\textbf{Arima}}                                                                       \\ \hline
\multicolumn{1}{|l|}{\textbf{Dataset}}    & \textbf{Model}           & \multicolumn{1}{l|}{\textbf{sMAPE}} & \multicolumn{1}{l|}{\textbf{NMAE}} & \multicolumn{1}{l|}{\textbf{NRMSE}} & \multicolumn{1}{l|}{\textbf{sMAPE}} & \multicolumn{1}{l|}{\textbf{NMAE}} & \multicolumn{1}{l|}{\textbf{NRMSE}} & \multicolumn{1}{l|}{\textbf{sMAPE}} & \multicolumn{1}{l|}{\textbf{NMAE}} & \multicolumn{1}{l|}{\textbf{NRMSE}} & \multicolumn{1}{l|}{\textbf{sMAPE}} & \multicolumn{1}{l|}{\textbf{NMAE}} & \multicolumn{1}{l|}{\textbf{NRMSE}} & \multicolumn{1}{l|}{\textbf{sMAPE}} & \multicolumn{1}{l|}{\textbf{NMAE}} & \multicolumn{1}{l|}{\textbf{NRMSE}} \\ \hline
\multirow{5}{*}{\textbf{M1}}              & LLMTime + E (GPT4)       & \multicolumn{1}{r|}{0.47}           & \multicolumn{1}{r|}{\textbf{0.48}} & \textbf{0.48}                       & \multicolumn{1}{r|}{0.41}           & \multicolumn{1}{r|}{\textbf{0.47}} & \textbf{0.48}                       & \multicolumn{1}{r|}{\textbf{0.44}}  & \multicolumn{1}{r|}{\textbf{0.25}} & \textbf{0.20}                       & \multicolumn{1}{r|}{0.46}           & \multicolumn{1}{r|}{0.42}          & 0.40                                & \multicolumn{1}{r|}{0.44}           & \multicolumn{1}{r|}{\textbf{0.44}} & \textbf{0.43}                       \\ \cline{2-17} 
                                          & LLMTime + E (Llama3-70b) & \multicolumn{1}{r|}{0.45}           & \multicolumn{1}{r|}{0.63}          & 0.67                                & \multicolumn{1}{r|}{0.44}           & \multicolumn{1}{r|}{0.49}          & 0.49                                & \multicolumn{1}{r|}{0.46}           & \multicolumn{1}{r|}{0.49}          & 0.49                                & \multicolumn{1}{r|}{\textbf{0.37}}  & \multicolumn{1}{r|}{0.42}          & 0.42                                & \multicolumn{1}{r|}{\textbf{0.40}}  & \multicolumn{1}{r|}{0.46}          & 0.44                                \\ \cline{2-17} 
                                          & LLMTime + E (Llama2-70b) & \multicolumn{1}{r|}{0.53}           & \multicolumn{1}{r|}{0.54}          & 0.55                                & \multicolumn{1}{r|}{0.50}           & \multicolumn{1}{r|}{0.51}          & 0.50                                & \multicolumn{1}{r|}{0.52}           & \multicolumn{1}{r|}{0.48}          & 0.47                                & \multicolumn{1}{r|}{0.49}           & \multicolumn{1}{r|}{0.51}          & 0.50                                & \multicolumn{1}{r|}{0.48}           & \multicolumn{1}{r|}{0.46}          & 0.45                                \\ \cline{2-17} 
                                          & LLMTime + E (Vicuna-7b)  & \multicolumn{1}{r|}{0.48}           & \multicolumn{1}{r|}{0.40}          & 0.42                                & \multicolumn{1}{r|}{0.48}           & \multicolumn{1}{r|}{0.56}          & 0.56                                & \multicolumn{1}{r|}{0.50}           & \multicolumn{1}{r|}{0.41}          & 0.41                                & \multicolumn{1}{r|}{0.44}           & \multicolumn{1}{r|}{\textbf{0.32}} & \textbf{0.31}                       & \multicolumn{1}{r|}{0.49}           & \multicolumn{1}{r|}{0.77}          & 0.74                                \\ \cline{2-17} 
                                          & LLMTime + E (Mistral-7b) & \multicolumn{1}{r|}{\textbf{0.39}}  & \multicolumn{1}{r|}{0.49}          & 0.49                                & \multicolumn{1}{r|}{\textbf{0.40}}  & \multicolumn{1}{r|}{0.59}          & 0.59                                & \multicolumn{1}{r|}{0.47}           & \multicolumn{1}{r|}{0.49}          & 0.50                                & \multicolumn{1}{r|}{0.46}           & \multicolumn{1}{r|}{0.63}          & 0.62                                & \multicolumn{1}{r|}{0.47}           & \multicolumn{1}{r|}{0.51}          & 0.52                                \\ \hline
\multicolumn{1}{|l|}{\multirow{5}{*}{M3}} & LLMTime + E (GPT4)       & \multicolumn{1}{r|}{\textbf{0.42}}  & \multicolumn{1}{r|}{0.49}          & 0.49                                & \multicolumn{1}{r|}{\textbf{0.44}}  & \multicolumn{1}{r|}{0.47}          & 0.47                                & \multicolumn{1}{r|}{\textbf{0.45}}  & \multicolumn{1}{r|}{\textbf{0.17}} & \textbf{0.14}                       & \multicolumn{1}{r|}{\textbf{0.42}}  & \multicolumn{1}{r|}{\textbf{0.40}} & 0.42                                & \multicolumn{1}{r|}{0.46}           & \multicolumn{1}{r|}{0.50}          & 0.49                                \\ \cline{2-17} 
\multicolumn{1}{|l|}{}                    & LLMTime + E (Llama3-70b) & \multicolumn{1}{r|}{0.42}           & \multicolumn{1}{r|}{\textbf{0.40}} & \textbf{0.40}                       & \multicolumn{1}{r|}{\textbf{0.44}}  & \multicolumn{1}{r|}{\textbf{0.35}} & \textbf{0.33}                       & \multicolumn{1}{r|}{0.46}  & \multicolumn{1}{r|}{0.48} & 0.48                       & \multicolumn{1}{r|}{0.45}           & \multicolumn{1}{r|}{0.41} & \textbf{0.41}                       & \multicolumn{1}{r|}{\textbf{0.42}}  & \multicolumn{1}{r|}{\textbf{0.39}} & \textbf{0.40}                       \\ \cline{2-17} 
\multicolumn{1}{|l|}{}                    & LLMTime + E (Llama2-70b) & \multicolumn{1}{r|}{0.52}           & \multicolumn{1}{r|}{0.48}          & 0.48                                & \multicolumn{1}{r|}{0.49}           & \multicolumn{1}{r|}{0.39}          & 0.40                                & \multicolumn{1}{r|}{0.49}           & \multicolumn{1}{r|}{0.64}          & 0.72                                & \multicolumn{1}{r|}{0.50}           & \multicolumn{1}{r|}{0.59}          & 0.58                                & \multicolumn{1}{r|}{0.49}           & \multicolumn{1}{r|}{0.44}          & 0.43                                \\ \cline{2-17} 
\multicolumn{1}{|l|}{}                    & LLMTime + E (Vicuna-7b)  & \multicolumn{1}{r|}{0.48}           & \multicolumn{1}{r|}{0.49}          & 0.48                                & \multicolumn{1}{r|}{0.47}           & \multicolumn{1}{r|}{0.50}          & 0.50                                & \multicolumn{1}{r|}{0.46}  & \multicolumn{1}{r|}{0.95}          & 0.96                                & \multicolumn{1}{r|}{\textbf{0.44}}  & \multicolumn{1}{r|}{0.50}          & 0.51                                & \multicolumn{1}{r|}{0.43}           & \multicolumn{1}{r|}{0.53}          & 0.52                                \\ \cline{2-17} 
\multicolumn{1}{|l|}{}                    & LLMTime + E (Mistral-7b) & \multicolumn{1}{r|}{0.44}           & \multicolumn{1}{r|}{0.60}          & 0.59                                & \multicolumn{1}{r|}{0.46}           & \multicolumn{1}{r|}{0.43}          & 0.44                                & \multicolumn{1}{r|}{0.46}  & \multicolumn{1}{r|}{0.81}          & 0.81                                & \multicolumn{1}{r|}{0.46}           & \multicolumn{1}{r|}{0.60}          & 0.59                                & \multicolumn{1}{r|}{0.51}           & \multicolumn{1}{r|}{0.64}          & 0.63                                \\ \hline
\multirow{5}{*}{\textbf{Tourism}}         & LLMTime + E (GPT4)       & \multicolumn{1}{r|}{\textbf{0.42}}  & \multicolumn{1}{r|}{\textbf{0.40}} & \textbf{0.39}                       & \multicolumn{1}{r|}{\textbf{0.43}}  & \multicolumn{1}{r|}{\textbf{0.30}} & \textbf{0.29}                       & \multicolumn{1}{r|}{\textbf{0.44}}  & \multicolumn{1}{r|}{0.62}          & 0.60                                & \multicolumn{1}{r|}{\textbf{0.41}}  & \multicolumn{1}{r|}{0.52}          & 0.50                                & \multicolumn{1}{r|}{\textbf{0.45}}  & \multicolumn{1}{r|}{\textbf{0.43}} & \textbf{0.43}                       \\ \cline{2-17} 
                                          & LLMTime + E (Llama3-70b) & \multicolumn{1}{r|}{0.43}           & \multicolumn{1}{r|}{0.45}          & 0.45                                & \multicolumn{1}{r|}{0.45}           & \multicolumn{1}{r|}{0.47}          & 0.48                                & \multicolumn{1}{r|}{\textbf{0.44}}  & \multicolumn{1}{r|}{\textbf{0.47}} & \textbf{0.47}                       & \multicolumn{1}{r|}{0.43}           & \multicolumn{1}{r|}{\textbf{0.42}} & \textbf{0.42}                       & \multicolumn{1}{r|}{0.46}           & \multicolumn{1}{r|}{0.46}          & 0.45                                \\ \cline{2-17} 
                                          & LLMTime + E (Llama2-70b) & \multicolumn{1}{r|}{0.49}           & \multicolumn{1}{r|}{0.49}          & 0.49                                & \multicolumn{1}{r|}{0.52}           & \multicolumn{1}{r|}{0.54}          & 0.54                                & \multicolumn{1}{r|}{0.48}           & \multicolumn{1}{r|}{0.73}          & 0.72                                & \multicolumn{1}{r|}{0.50}           & \multicolumn{1}{r|}{0.48}          & 0.48                                & \multicolumn{1}{r|}{0.50}           & \multicolumn{1}{r|}{0.80}          & 0.78                                \\ \cline{2-17} 
                                          & LLMTime + E (Vicuna-7b)  & \multicolumn{1}{r|}{0.48}           & \multicolumn{1}{r|}{0.48}          & 0.48                                & \multicolumn{1}{r|}{0.50}           & \multicolumn{1}{r|}{0.48}          & 0.48                                & \multicolumn{1}{r|}{0.50}           & \multicolumn{1}{r|}{0.48}          & 0.48                                & \multicolumn{1}{r|}{0.43}           & \multicolumn{1}{r|}{0.97}          & 0.97                                & \multicolumn{1}{r|}{0.48}           & \multicolumn{1}{r|}{0.61}          & 0.59                                \\ \cline{2-17} 
                                          & LLMTime + E (Mistral-7b) & \multicolumn{1}{r|}{0.44}           & \multicolumn{1}{r|}{0.99}          & 0.99                                & \multicolumn{1}{r|}{0.45}           & \multicolumn{1}{r|}{0.68}          & 0.69                                & \multicolumn{1}{r|}{0.46}           & \multicolumn{1}{r|}{0.88}          & 0.87                                & \multicolumn{1}{r|}{0.44}           & \multicolumn{1}{r|}{0.51}          & 0.51                                & \multicolumn{1}{r|}{0.47}           & \multicolumn{1}{r|}{0.76}          & 0.74                                \\ \hline
\end{tabular}
}
\caption{Performance comparison across different LLMs as baseline explainer backbone for synthetic simulatability. We report NMAE and NRMSE additional to sMAPE
from the main paper results, lower is better.}
\label{tab_app:in_sim_llms}
\end{table*}

\end{document}